\documentclass[runningheads]{llncs}

\usepackage[year=2026]{eccv}

\usepackage{eccvabbrv}

\usepackage{graphicx}
\usepackage{booktabs}
\usepackage{multirow}
\usepackage{makecell}
\usepackage{tabularx}
\usepackage[table]{xcolor}
\usepackage[table]{xcolor}
\usepackage{makecell}
\usepackage{cuted}
\usepackage{caption}
\usepackage{subcaption}
\usepackage{xcolor}

\usepackage{microtype}
\usepackage{nicefrac}
\usepackage[american]{babel}
\usepackage{multicol}
\usepackage{wrapfig}

\usepackage[accsupp]{axessibility}  
\usepackage[a4paper, left=1.45in, right=1.45in, top=1.4in, bottom=1.4in]{geometry}

\ifodd 0
\newcommand{\rev}[1]{{\color{blue}#1}} 
\else
\newcommand{\rev}[1]{#1}
\fi
\usepackage[pagebackref,breaklinks,colorlinks,citecolor=eccvblue]{hyperref}

\usepackage{orcidlink}
\newcommand{\name}{ZoomUI\xspace}

\begin{document}

\title{Zoom to Essence: Trainless GUI Grounding by Inferring upon Interface Elements} 

\titlerunning{Zoom to Essence}

\author{Ziwei Liu\inst{1} \and
Tao Feng\inst{2} \and
Borui Kang\inst{3} \and
Yanbing Yang\inst{1} \and
Jun Luo$^\dagger$\inst{4}}

\authorrunning{Liu. et al.}

\institute{Department of Computer Science and Technology, Sichuan University, China \and
Department of Computer Science and Technology, Tsinghua University, China \and
School of Computer Science, Nanjing University, China \and
College of Computing and Data Science, Nanyang Technological University, Singapore
}

\maketitle

\begin{abstract}
Multimodal Large Language Model (MLLM)-based Graphical User Interface (GUI) agents develop rapidly, with \textit{visual grounding} that maps natural language instructions to target UI elements serving as the core capability. Existing GUI agents typically fine-tune MLLM on massive datasets to handle challenges in understanding instructions and UI interfaces, which not only incurs high data annotation costs but also makes performance dependent on data quality and distribution. To avoid such cumbersome yet ineffective training, we notice that complex UI interfaces can be decomposed into basic visual elements directly understandable by common MLLMs. Consequently, we propose \name that leverages inference scaling to guide common MLLMs in progressively anchor instruction elements to increasingly detailed interface elements.
Specifically, \name first optimizes the latent thinking to transform original instruction into element visual features description, and subsequently leverages internal attention to iteratively zoom in target element interface region. Evaluations on extensive benchmarks demonstrate that \name reaches or even surpasses SOTA baselines.
\end{abstract}

\section{Introduction}
\label{sec:intro}
Emerging as a novel interaction paradigm, autonomous GUI agents powered by MLLM can interpret users' instructions to execute tasks—such as clicking icons or localizing elements—across complex digital applications~\cite{cogagent,minicpm,internvl,claude,guiaif}. The cornerstone of this capability is \textit{visual grounding}, which maps natural language instructions to precise pixel-level spatial coordinates of target element on the interactive UI interfaces~\cite{gui-g1,guiactor,guig2,infigui,ui-ins,uitars,uivenus}. 
Although 
GUI grounding 
appears to be 
promising 
in simple UI environments~\cite{seeclick,osatlas}, they still severely struggle in scenarios characterized by ambiguous user instructions and information-dense high resolution interfaces~\cite{screenspotpro,uivision}, as these challenges often prevent GUI agents from precisely identifying the spatial locations of target elements.

Existing GUI agents typically resolve these challenges by fine-tuning MLLM on massive offline datasets~\cite{uground,gta1,guiactor,opencua}, but such data-driven methods inevitably face their limitations.
Firstly, constructing such massive GUI datasets typically requires extensive annotation either manually or via advanced proprietary models (e.g., ChatGPT and Gemini)~\cite{mind2web,omniact,showui,widget}, both of which incur high costs. 
Secondly, these datasets are inherently plagued by errors and noisy annotations, 
so the grounding performance of the fine-tuned GUI agents is made to be significantly relying on training data quality and distribution. 
Consequently, a question naturally arises: \textit{Is it possible to tackle instruction and interface challenges without relying on any agent fine-tuning?}

\begin{figure*}[t!]
    \centering
    \includegraphics[width=0.99\linewidth]{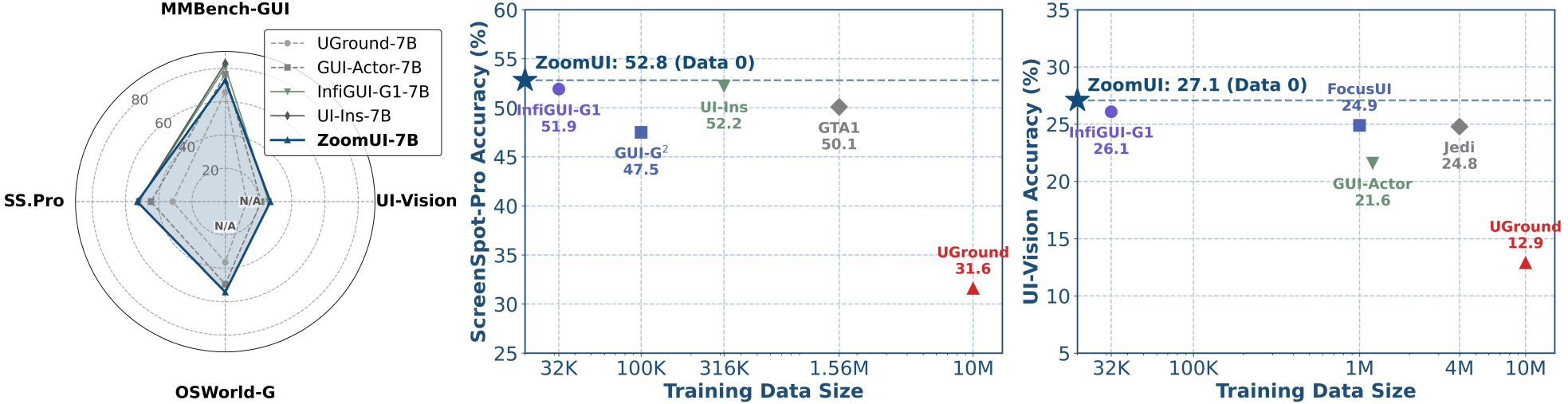}
    \caption{Performance comparisons of \name and other SOTA baseline methods.
    }
    \label{fig:zoomui}
    \vspace{-2ex}
\end{figure*}


In fact, fine-tuning MLLM may not be necessary upon challenging scenarios. Let us consider an instruction ``clear the search'' issued upon a browser interface: whereas GUI agents should click the small `$\times$' icon adjacent to search bar, 
%
%
they are often confused by this over-simplified instruction and thus try to directly map it to certain graphical elements (or even the texts) contained in the fine-tuning dataset.
%
%
Fortunately, 
most MLLMs can readily decompose the instruction into two semantic tokens: ``search'' and ``clear'', and they might respectively guide an MLLM (e.g., \cite{qwen2.5}) to locate the search bar region and the `$\times$' icon adjacent to it, hence achieving GUI grounding in a zero-shot manner.
Of course, scenarios faced by GUI agents cannot be so straightforward in reality. 
On one hand, keyword like ``clear'' may introduce semantic ambiguity, potentially misguiding the MLLM toward distractors elsewhere on the interface, such as a ``recycle bin'' icon.
On the other hand, the target `$\times$' icon typically occupies a negligible pixel area. Within an information dense and high-resolution interface, such features are easily overwhelmed by unrelated elements, hence exceeding the perception limits of MLLM's vision encoder.
%


To this end, we propose \name
that leverages inference scaling on a general MLLM to decompose GUI grounding into basic element understanding without data-intensive fine-tuning.
Specifically, \name decomposes the inference process into two steps.
It first refines the ambiguous instruction by guiding MLLM to generate visual features description of the target element on the UI interface. 
This is achieved by injecting learnable thought vectors into the MLLM's latent space and then optimizing these vectors 
to maximize the output generation likelihood.
This exploration process ensures the model possesses sufficient confidence to comprehensively consider related contextual cues (such as binding ``clear'' with ``search''). Consequently, it deduces the visual profile of the target element (e.g., the `$\times$' adjacent to search bar) rather than being misled by irrelevant ones.
Building upon this refined instruction, \name subsequently exploits
the MLLM's latent attention distribution as guidance during the element coordinate generation phase, so as to progressively zoom in on the specific area containing the target element. Evaluations conducted across multiple
benchmarks validate the performance of our approach, as illustrated in Figure~\ref{fig:zoomui}.
%
Overall, the main contributions are:

\begin{itemize}
 
    \item We propose \name, a training-free framework that leverages inference scaling to decompose GUI grounding tasks into basic visual element understanding, enables general MLLM's GUI grounding ability without data fine-tuning.

    \item We design two inference steps: a latent thought vectors optimization to refine ambiguous instructions into element visual features description; an iterative visual focus that leverages attention distribution during coordinate generation to zoom in target element region.
 
    \item Evaluations on extensive GUI benchmarks demonstrate that \name reaches or even surpasses SOTA baselines.

\end{itemize}

\section{Related Work}
\subsection{GUI Agents and Grounding}
Beyond brittle script-based paradigms, recent GUI agents utilize MLLM to achieve visual perception and natural language understanding for automatic digital human-computer interactions. Central to this capability is grounding, which locates UI elements based on instructions serving as a basic prerequisite and primary metric for GUI agents' performance. Representative early attempts, OmniParser~\cite{omniparser}, OS-Atlas~\cite{osatlas}, and SeeClick~\cite{seeclick} leverage UI interfaces and fine-tune on MLLM to reach GUI grounding in cross-platform UI environments. Reinforcement learning (RL) has rapidly emerged as a paradigm for advancing MLLM, and is leveraged to further enhance GUI grounding performance~\cite{segui,guir1,infigui,guig2}. They leverage group relative policy optimization (GRPO)~\cite{grpo} and explore efficient reward strategies to improve visual perception of GUI agents. For cross-platform diversity and instruction understanding bottleneck, GUI-AiF~\cite{guiaif} and UI-Ins~\cite{ui-ins} leverage refined RL techniques to fine-tune MLLM. Besides, FocusUI~\cite{focusui} properly reduces unrelated visual tokens of UI interface for grounding. However, despite their effectiveness, these methods rely heavily on computationally expensive fine-tuning using large-scale annotated UI datasets, neglecting the potential that general MLLMs can achieve GUI grounding by proper inference scaling.

\subsection{Inference Scaling}
Inference scaling refers to the paradigm of enhancing large language model (LLM) performance by increasing the budget at test time without adjusting model parameters, enabling models to solve tasks through iterative deliberation rather than immediate generation, which has been employed in mathematical reasoning~\cite{ismath,ismath2}, code generation~\cite{iscode}, and workflow planning~\cite{isplanning}. For instance, Chain-of-thought (CoT)~\cite{cot} that forces models to generate reasoning steps before gives the final answer, significantly improves model performance. Besides, best-of-N sampling generates multiple candidate solutions to select the optimal outcome~\cite{scalingllm}, and external verification leverages tools to validate the correctness of intermediate steps~\cite{revise}. Inspired by its success in LLM, some methods attempt similar techniques in GUI grounding. DiMo-GUI~\cite{dimogui} explicitly disentangles text and icon elements for separate inferring, yet it relies on the performance of auxiliary tools (e.g., OCR detectors). RegionFocus~\cite{regionfocus} employs slicing rules (e.g., fixed grids) for zooming into UI interface, while such rigid partitioning disregards MLLM's intrinsic understanding of the interface.

\begin{figure*}[t!]
    \centering
    \includegraphics[width=0.9\linewidth]{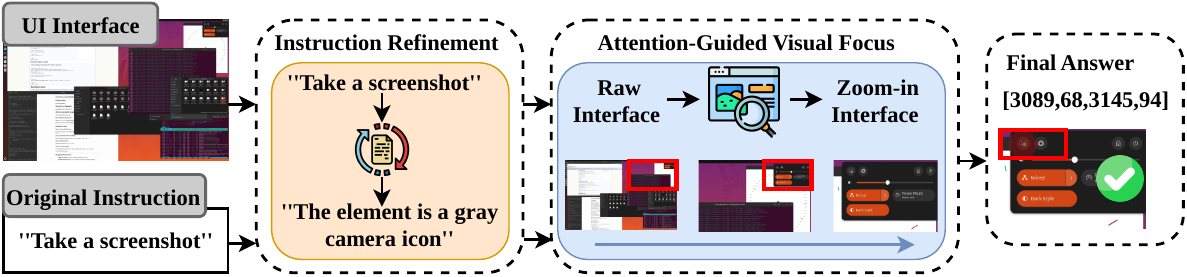}
    \caption{\textbf{Overview Workflow.} \name initiates by refining the original instruction to a visual features description of element. Subsequently, by capturing attention distribution during coordinate generation phase, it iteratively zooms into relevant regions to obtain more a fine-grained interface of target UI element.
    }
    \label{fig:overview}
    \vspace{-2ex}
\end{figure*}

\section{Method}

\subsection{Overview}
\label{overview}
The task of GUI grounding is to align a user’s
natural language instruction with its corresponding interactive element on a UI interface at the pixel level. Given an interface $\mathbf{s} \in \mathbb{R}^{H \times W \times 3}$ and an user instruction $\mathbf{i}$, the final goal is to predict a bounding box $\mathbf{b}=[x_{1},y_{1},x_{2},y_{2}]$ that precisely locates the target element, where $[x_{1},y_{1}]$ and $[x_{2},y_{2}]$ denote the top-left and bottom-right corners respectively. Our inference framework first refines the original instruction into a description regarding target element's visual features and then iteratively zooms in element region. Specifically, we convert an instruction $\mathbf{i}$ (e.g.,``take a screenshot'') into a visual description $\mathbf{i}_{\text{vis}}$ (e.g.,``a gray camera icon''), thereby bridging the semantic gap between user intent and observable features. Guided by this $\mathbf{i}_{\text{vis}}$, the framework then initiates an iterative visual focus process. By analyzing model's attention distribution during predicting element coordinates, we crop a region from the original interface and upscale it for zoom-in observation. In each step, the region is formed for the next iteration. These iterative procedures allow for reducing disturbance from unrelated elements and resolving high-resolution problems that limit model's visual perception, ultimately enhancing prediction performance of final bounding box $\mathbf{b}$. The workflow is shown in Figure~\ref{fig:overview}.

\begin{figure*}[t!]
    \centering
    \includegraphics[width=0.68\linewidth]{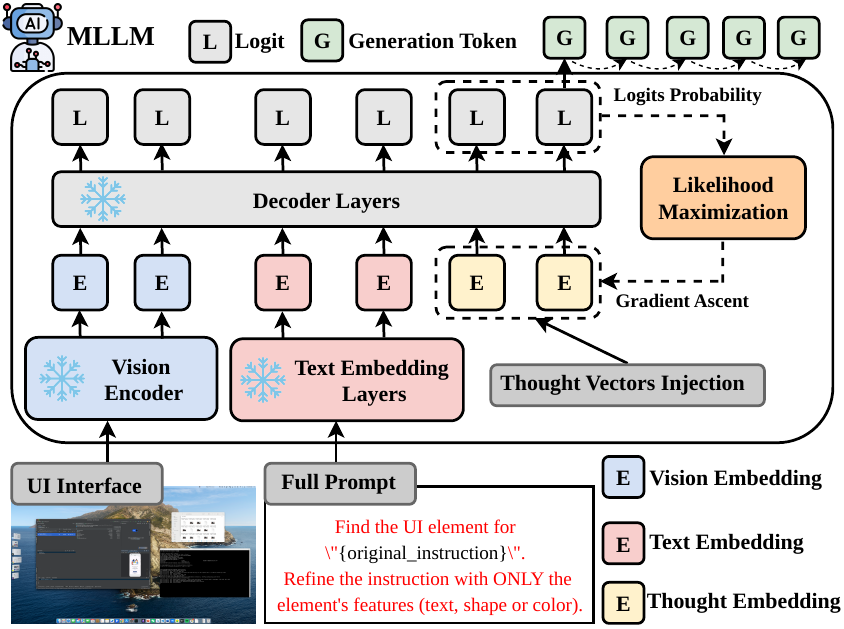}
    \caption{\textbf{Instruction Refinement.} We introduce learnable thought vectors injected in the interface and prompt embeddings. These vectors are iteratively optimized by maximizing the likelihood of the output logits via gradient ascent, which steers the latent representations to generate more reliable refined instructions.
    }
    \label{fig:ins}
    \vspace{-2ex}
\end{figure*}

\subsection{Instruction Refinement}
\label{ins}
To bridge the gap between instruction and specific element features, we leverage a latent thinking strategy that operates in model's latent space without adjusting parameters. Our objective is to activate MLLM's GUI element understanding by optimizing its internal thinking process, hence analyzing the interface and generate visual features (e.g., text, color, or shape) that match user's intent. The instruction refinement workflow is shown in Figure~\ref{fig:ins}.

\subsubsection{Instruction Reasoning Representations.} We construct a hybrid query embedded into model's standard inference flow. Let $\mathcal{V}(\cdot)$ and $\mathcal{E}(\cdot)$ denote the vision encoder and text embedding layers of the model, respectively. In a standard setting, model directly processes the concatenated sequence of interface embeddings $\mathcal{V}(\mathbf{s})$ and textual embeddings $\mathcal{E}(\mathbf{t})$ of instruction $\mathbf{i}$ within a task-specific prompt to generate a response. To enable model thinking before generating refined instruction, we inject a sequence of learnable latent thought vectors, denoted as $\mathbf{v} \in \mathbb{R}^{N \times D}$ into the end of input stream, where $N$ is the sequence length and $D$ is the embedding dimension. The total input sequence $X_{in}$ can be formulated as:
\begin{equation}
    X_{in} = [\mathcal{V}(\mathbf{s}), \mathcal{E}(\mathbf{t}), \mathbf{v}].
\end{equation}
Unlike the discrete tokens used in standard prompting, $\mathbf{v}$ act as a differentiable vectors. These vectors serve as active parameters during the instruction refining phase, initialized to a semantic starting point, but ready to be optimized via gradients to prompt the visual features of target element.

\subsubsection{Likelihood-based Optimization.} To guide the instruction reasoning optimization, we derive an intrinsic signal directly from the model's token output probability distribution. We assume the model's generation probability serves as a proxy for our targeted description. Specifically, when the model assigns high probability to a generated description, it indicates that the latent thought vectors more likely understand correct visual features of elements, thereby resolving original instruction's ambiguity. Conversely, low probability implies a hesitation to output a reliable description. Formally, we model this optimization process as a likelihood maximization problem. First, for each decoding step $t$, the model $\Theta$ analyzes the $X_{in}$ to output a probability distribution over the vocabulary $\mathcal{V}$:
\begin{equation}
      P_t(w) = \text{softmax}(\text{Logits}_\Theta(w \mid y_{<t}, X_{in})) \quad \text{for } w \in \mathcal{V},
\end{equation}
where $P_t(w)$ denotes the predicted probability of a candidate token $w$ at decoding step $t$, conditioned on the sequence of previously generated tokens $y_{<t} = (y_1, \dots, y_{t-1})$. 

Subsequently, to evaluate the reliability of token $w$, at this step, we define the confidence score as its log-probability:
\begin{equation}
    C_t(\mathbf{v}) = \log P_t(w).
\end{equation}
A higher score of $C_t(\mathbf{v})$ indicates model is more confidential for this token given the UI visual interface, maximizing this score encourages the latent thought vectors to reason more accurate and unambiguous visual description. Finally, to ensure global consistency of the description, we aggregate all $C_t(\mathbf{v})$ across the generated sequence of length $M$. The global optimization objective $\mathcal{J}(\mathbf{v})$ is defined as the average confidence:
\begin{equation} 
    \mathcal{J}(\mathbf{v}) = \frac{1}{M} \sum_{t=1}^{M} C_t(\mathbf{v}). 
\end{equation}

\subsubsection{Thought Vectors Gradient Update.}
At each step $k$, we perform a differentiable forward pass to compute the objective $\mathcal{J}(\mathbf{v}_k)$. By leveraging backpropagation to obtain the exact gradient $\nabla_{\mathbf{v}} \mathcal{J}(\mathbf{v}_k)$, we update latent thought vectors via gradient ascent:
\begin{equation} 
    \mathbf{v}_{k+1} = \mathbf{v}_k + \eta \cdot \nabla_{\mathbf{v}} \mathcal{J}(\mathbf{v}_k), 
\end{equation}
where $\eta$ is the learning rate. This iterative refinement steers the latent thoughts to maximize the model's certainty, ensuring the final description $\mathbf{i}_{\text{vis}}$ is concisely grounded in the visual evidence. \rev{We will further discuss the effects of these parameters in \textbf{Appendix A}.}

\begin{figure*}[t!]
    \centering
    \includegraphics[width=0.99\linewidth]{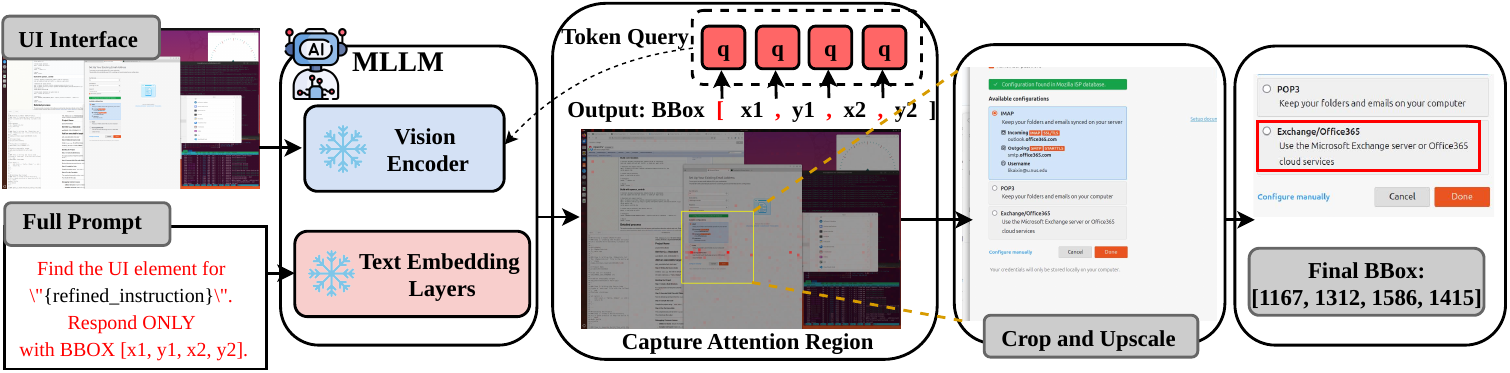}
    \caption{\textbf{Attention-Guided Visual Focus.} We capture the attention distribution on the UI interface during the coordinate prediction phase. \rev{By iteratively cropping and upscaling a region with the highest attention scores, we progressively narrow target element region to enhance final grounding performance.}
    }
    \label{fig:visualfocus}
    \vspace{-2ex}
\end{figure*}

\subsection{Attention-Guided Visual Focus}
\label{visual}
Instruction refinement obtains a visual features description about target element, while MLLM is susceptible to the influence of high-resolution UI interface during locating precise click coordinate $\mathbf{b}=[x_ {1}, y_ {1}, x_ {2}, y_ {2}]$. \rev{As illustrated in Figure~\ref{fig:visualfocus}, the ground-truth target might occupy merely a few dozen pixels within a 4K interface. This extreme scale disparity necessitates an attention-guided mechanism to iteratively zoom into the observation region for fine-grained visual details, thereby enhancing the final grounding.}

\subsubsection{Attention Region Capture.}
Instead of directly generating coordinate $\mathbf{b}$ based on refined instruction, we first investigate model's visual attention distribution on UI interface during element coordinate generation phase. \rev{We argue that the model has established a deeply understanding from refined instruction to interface at this stage.
Therefore, its attention should represent the most faithful grounding of the target element's position.} Furthermore, since we leverage $\mathbf{b}=[x_{1}, y_{1}, x_{2}, y_{2}]$ as the output format, to investigate the model's attention prior to predicting each spatial coordinate, we employ the opening bracket (``['') and the commas (``,'') as probing queries to hook the corresponding visual attention distribution. Formally, let $\mathcal{T} = \{t_0, t_1, t_2, t_3\}$ denote the probing tokens in the generated sequence. For the selected decoder layer $l$ and attention head $h$, the cross-attention map $A^{(l)}_{h, t} \in \mathbb{R}^{H \times W}$ at step $t \in \mathcal{T}$ is computed by querying the visual features $K_\mathcal{V}^{(l)}$ with the hidden state $q_t^{(l)}$ of the probing token:
\begin{equation}
    A^{(l)}_{h, t} = \text{Softmax}\left(\frac{q_t^{(l)} (K_\mathcal{V}^{(l)})^\top}{\sqrt{d_k}}\right),
\end{equation}
where $d_k$ is the dimension of the key vectors. To extract the dominant visual cues while filtering out noise, we employ a max-pooling strategy. We first aggregate attention heads to identify the most confident attention region, and then across the trigger steps to encompass the element's full spatial extent. The final attention map $M$ is derived from the selected layer $l$:
\begin{equation}
    M = \max_{t \in \mathcal{T}, \, h} A^{(l)}_{h, t}.
\end{equation}

\subsubsection{Iterative Zoom-in Processing.}
Given the $M$ derived from the attention distribution, we process to the iterative visual focus phase. To narrow the grounding region, we define a fixed zoom-in window size denoted as $(H_{z}, W_{z})$. We argue that the region containing the target element exhibits the highest scores of attention. Mathematically, we quantify attention scores using a sliding window approach. Given the patch size $P$, we first map the window dimensions to the feature grid scale as $(h_z, w_z) = (H_z/P, W_z/P)$. To locate this region, we compute the maximum accumulated scores $S(u, v)$ by spatially aggregating attention values within a sliding window of grid size $(h_z, w_z)$:
\begin{equation}
    S(u, v) = \sum_{i=0}^{h_z-1} \sum_{j=0}^{w_z-1} M(u+i, v+j).
\end{equation}
The center point $(u^*, v^*)$ is then obtained by locating the peak of this map:
\begin{equation}
    (u^*, v^*) = \arg\max_{u, v} S(u, v).
\end{equation}
We then project this center back to the UI interface coordinate space to obtain the cropping center $(c_x, c_y)$. Using the center and window size, we crop the region from the original interface, and then employ bicubic interpolation to upscale elements in the cropped interface. We iterate crop and upscale cycle, using the cropped region to guide next round of attention analysis, thereby progressively focusing the target element region until iteration count is reached.

\section{Experiments and Results}
\subsection{Experimental Setup}
\subsubsection{Baselines and Metrics.} We compare our method’s grounding performance against extensive GUI agent baselines. These include proprietary models such as GPT-4o~\cite{gpt4.1}, Claude~\cite{claude} and Gemini-2.5-Pro~\cite{gemini}. For fine-tuning based GUI agents, we choose ‌representative methods adopting supervised fine-tuning and RL paradigm, which includes InfiGUI-R1~\cite{infigui}, GTA1~\cite{gta1}, SE-GUI~\cite{segui}, GUI-Actor~\cite{guiactor}, GUI-G$^2$~\cite{guig2}, UI-Ins~\cite{ui-ins}, FocusUI~\cite{focusui} and so on.

Following prior works, we evaluate GUI
grounding performance using the point-in-box accuracy. Specifically, given that the model predicts a bounding box denoted as $\hat{b} = (\hat{x}_{1}, \hat{y}_{1}, \hat{x}_{2}, \hat{y}_{2})$, we first derive its geometric centroid $p_c = (\frac{\hat{x}_{1} + \hat{x}_{2}}{2}, \frac{\hat{y}_{1} + \hat{y}_{2}}{2})$. A prediction is deemed valid only if the derived center $p_c$ resides within the spatial confines of the ground-truth box $b_{gt}$. Consequently, the overall accuracy across $N$ test samples is formalized as:
\begin{equation}
    \text{Accuracy} = \frac{1}{N} \sum_{i=1}^{N} \left( p_{c}^{(i)} \in \text{Area}(b_{gt}^{(i)}) \right).
\end{equation}

\subsubsection{Implementation Details.} To ensure a fair comparison with baselines, we employ Qwen2.5-VL-3B-Instruct and Qwen2.5-VL-7B-Instruct as our backbone MLLM architectures, \rev{that are matched with baseline methods}. We do not train on any UI dataset and fine-tune any parameters of backbone MLLM, relying solely on the proposed inference scaling method. All experiments are conducted on one H100 80G GPU. For default configuration, we extract attention distribution during generation from the decoder layer at the 70~\% depth. The window size is fixed at 784x784. For a maximum of 3 iterations, the visual focus region is cropped to 1/2 and subsequently upscaled by 2x via pixel interpolation, ensuring the input window size remains constant.

\subsubsection{Evaluation Benchmarks.} 
We evaluate \name on four representative GUI benchmarks, each chosen to assess distinct capabilities. \textbf{ScreenSpot-Pro}~\cite{screenspotpro} is the evaluation of GUI grounding in professional software applications, utilizing ultra-high resolution interfaces (up to 4K and 6K) to simulate realistic, information-dense workspaces. \textbf{OSWorld-G}~\cite{jedi} is built based on interfaces of real operating systems. Furthermore, this benchmark specifically introduces a rejection capability assessment to clarify whether GUI agents can correctly reject generating a bounding box when an element described by an instruction does not appear in the interface.
\textbf{UI-Vision}~\cite{uivision} is designed to evaluate generalization across a broad spectrum of desktop applications. By classifying cases based on diverse element types and informative layout structures, it provides a granular assessment of GUI agents' performance within unseen UI environments. \textbf{MMbench-GUI} covers comprehensive digital interfaces-including mobile (Android, iOS), desktop (Windows, macOS, Linux), and web environments, which is designed to evaluate complex cross-platform performance of GUI agents.

\subsection{Main Results}
We present the main results of our evaluations in Table~\ref{sspro},~\ref{osworld},~\ref{uivision} and~\ref{mmbench}. The results consistently show that our \name reaches or even establishes new SOTA performance among open-source models in both the 3B and 7B parameter categories. Notably, our method also exhibits superior performance against several proprietary models with significantly larger parameter counts, highlighting the efficacy and efficiency of our proposed framework.

Our method demonstrates strong generalization capability by achieving consistently high performance across multiple benchmarks. \rev{Rather than fitting on specific UI training data, \name dynamically decomposes unseen and complex GUI grounding tasks into basic steps. This allows MLLM to adaptively interpret novel interfaces based on inherent knowledge rather than trained parameters, yielding highly generalized performance.}


\begin{table}[t!]
\centering
\caption{Performance comparison on the \textbf{ScreenSpot-Pro} benchmark. \textbf{Bold} and \underline{underline} represents the best and second results, respectively.}
\resizebox{0.99\textwidth}{!}{
    \setlength{\tabcolsep}{3pt} 
    \begin{tabular}{l | c | cc cc cc cc cc cc | c}
    \toprule
    \multirow{2}{*}{\textbf{Model}} & \multirow{2}{*}{\makecell{\textbf{Data}\\\textbf{Size}}} & \multicolumn{2}{c}{\textbf{CAD}} & \multicolumn{2}{c}{\textbf{Dev.}} & \multicolumn{2}{c}{\textbf{Creative}} & \multicolumn{2}{c}{\textbf{Scientific}} & \multicolumn{2}{c}{\textbf{Office}} & \multicolumn{2}{c|}{\textbf{OS}} & \multirow{2}{*}{\textbf{Avg.}} \\
    \cmidrule(lr){3-4} \cmidrule(lr){5-6} \cmidrule(lr){7-8} \cmidrule(lr){9-10} \cmidrule(lr){11-12} \cmidrule(lr){13-14}
     &  & Text & Icon & Text & Icon & Text & Icon & Text & Icon & Text & Icon & Text & Icon & \\
    \midrule
    \rowcolor{gray!20} 
    \multicolumn{15}{l}{\textit{Proprietary Model}}\\
    GPT-4o~\cite{gpt4.1} & - & 2.0 & 0.0 & 1.3 & 0.0 & 1.0 & 0.0 & 2.1 & 0.0 & 1.1 & 0.0 & 0.0 & 0.0 & 0.8 \\
    Claude C.~\cite{claude} & - & 14.5 & 3.7 & 22.0 & 3.9 & 25.9 & 3.4 & 33.9 & 15.8 & 30.1 & 16.3 & 11.0 & 4.5 & 17.1 \\
    \midrule
    \rowcolor{gray!20} 
    \multicolumn{15}{l}{\textit{Fine-Tuning based GUI Agents}}\\
    Qwen2.5-VL-3B~\cite{qwen2.5} & - & 9.1 & 7.3 & 22.1 & 1.4 & 26.8 & 2.1 & 38.2 & 7.3 & 33.9 & 15.1 & 10.3 & 1.1 & 16.1 \\
    UI-R1-3B~\cite{uir1} & 3K & 11.2 & 6.3 & 22.7 & 4.1 & 27.3 & 3.5 & 42.4 & 11.8 & 32.2 & 11.3 & 13.1 & 4.5 & 17.8 \\
    ZonUI-3B~\cite{zonui} & 24K & 31.9 & \underline{15.6} & 24.6 & 6.2 & 40.9 & 7.6 & 54.8 & 18.1 & 57.0 & 26.4 & 19.6 & 7.8 & 28.7 \\
    InfiGUI-R1-3B~\cite{infigui} & 32K & 33.0 & 14.1 & 51.3 & 12.4 & 44.9 & 7.0 & \underline{58.3} & 20.0 & 65.5 & 28.3 & \underline{43.9} & \underline{12.4} & 35.7 \\
    GUI-G1-3B~\cite{gui-g1}& 17K & \underline{39.6} & 9.4 & 50.7 & 10.3 & 36.6 & \underline{11.9} & \textbf{61.8} & \textbf{30.0} & \textbf{67.2} & \underline{32.1} & 23.5 & 10.6 & 37.1 \\
    SE-GUI-3B~\cite{segui}& 3K & 38.1 & 12.5 & \underline{55.8} & 7.6 & \underline{47.0} & 4.9 & \textbf{61.8} & 16.4 & 59.9 & 24.5 & 40.2 & \underline{12.4} & 35.9 \\
    JEDI-3B~\cite{jedi}& 4M & 27.4 & 9.4 & \textbf{61.0} & \underline{13.8} & \textbf{53.5} & 8.4 & 54.2 & 18.2 & 64.4 & \underline{32.1} & 38.3 & 9.0 & 36.1 \\
    FocusUI-3B~\cite{focusui}& 1M & - & - & - & - & - & - & - & - & - & - & - & - & \textbf{43.8} \\
    \rowcolor[HTML]{E8F0F8} \textbf{\name-3B} & \textbf{0} & \textbf{47.7} & \textbf{20.3} & 52.6 & \textbf{18.6} & 45.5 & \textbf{14.0} & 52.8 & \underline{25.5} & \underline{62.2} & \textbf{37.7} & \textbf{54.2} & \textbf{22.5}  & \underline{40.2} \\
    \midrule
    \rowcolor{gray!20} 
    \multicolumn{15}{l}{\textit{Fine-Tuning based GUI Agents}}\\
    Qwen2.5-VL-7B~\cite{qwen2.5} & - & 16.8 & 1.6 & 46.8 & 4.1 & 35.9 & 7.7 & 49.3 & 7.3 & 52.5 & 20.8 & 37.4 & 3.8 & 26.8 \\
    GUI-R1-7B~\cite{guir1} & 3K & 23.9 & 6.3 & 49.4 & 4.8 & 38.9 & 8.4 & 55.6 & 11.8 & 58.7 & 26.4 & 42.1 & 16.9 & 31.0 \\
    UGround-7B~\cite{uground} & 10M & 51.3 & 5.5 & 48.5 & 8.3 & 18.8 & 1.6 & 59.7 & 14.6 & 59.9 & 17.0 & 40.2 & 7.9 & 31.6 \\
    UI-TARS-7B~\cite{uitars} & - & 20.8 & 9.4 & 58.4 & 12.4 & 50.0 & 9.1 & 63.9 & \textbf{31.8} & 63.3 & 20.8 & 30.8 & 16.9 & 35.7 \\
    UI-AGILE-7B~\cite{uiagile} & 9K & 49.2 & 14.1 & 64.3 & 15.2 & 53.0 & 9.8 & 72.9 & 25.5 & 75.1 & 30.2 & 45.8 & 20.2 & 44.0 \\
    GUI-Actor-7B~\cite{guiactor} & 1.2M & 47.7 & 9.4 & 59.1 & 15.9 & 59.6 & 16.1 & 70.1 & 25.5 & 69.5 & 41.5 & 55.1 & 19.1 & 44.6 \\
    SE-GUI-7B~\cite{segui} & 3K & 51.3 & 14.1 & 68.2 & 19.3 & 57.6 & 9.1 & 75.0 & 28.2 & 78.5 & \underline{43.4} & 49.5 & 25.8 & 47.2 \\
    GUI-$G^2$-7B~\cite{guig2} & 100K & 55.8 & 12.5 & 68.8 & 17.2 & 57.1 & 15.4 & 77.1 & 24.5 & 74.0 & 32.7 & \underline{57.9} & 21.3 & 47.5 \\
    FocusUI-7B~\cite{focusui}& 1M & - & - & - & - & - & - & - & - & - & - & - & - & 48.3 \\
    OpenCUA-7B~\cite{opencua} & 35B & - & - & - & - & - & - & - & - & - & - & - & - & 50.0 \\
    GTA1-7B~\cite{gta1} & 1.56M & 53.3 & 17.2 & 66.9 & 20.7 & 62.6 & \underline{18.9} & 76.4 & \textbf{31.8} & \textbf{82.5} & \textbf{50.9} & 48.6 & 25.9 & 50.1 \\
    UI-Venus-7B~\cite{uivenus} & 107K & \underline{60.4} & 21.9 & \underline{74.7} & \underline{24.1} & 63.1 & 14.7 & 76.4 & \textbf{31.8} & 75.7 & 41.5 & 49.5 & 22.5 & 50.8 \\
    InfiGUI-G1-7B~\cite{infigui} & 32K & 57.4 & \underline{23.4} & \underline{74.7} & \underline{24.1} & \underline{64.6} & 18.2 & \underline{80.6} & \textbf{31.8} & 75.7 & 39.6 & 57.0 & \underline{29.2} & 51.9 \\
    UI-Ins-7B~\cite{ui-ins} & 316K & \textbf{60.9} & 20.3 & \textbf{75.3} & 18.6 & \textbf{65.2} & \underline{18.9} & \textbf{81.3} & \underline{29.1} & 79.7 & 37.7 & 57.0 & 25.8 & \underline{52.2} \\
    \midrule
    \rowcolor[HTML]{E8F0F8} \textbf{\name-7B} &  \textbf{0} & 59.9 & \textbf{25.0} & 64.9 & \textbf{30.3} & 56.6 & \textbf{23.1} & 67.4 & \textbf{31.8} & \underline{81.4} & \underline{43.4} & \textbf{67.3} & \textbf{34.8} & \textbf{52.8} \\
    \bottomrule
    \end{tabular}
}
\label{sspro}
\vspace{-2ex}
\end{table}

\begin{table}[t!]
\centering
\caption{Performance comparison on \textbf{OSWorld-G}. \textbf{Bold} and
\underline{underline} represents the best and second results, respectively.}
\resizebox{0.84\textwidth}{!}{
    \setlength{\tabcolsep}{5pt}
    \begin{tabular}{l | c | ccccc | c}
    \toprule
    \textbf{Model} & \textbf{Data Size} & \textbf{Text} & \textbf{Elem} & \textbf{Layout} & \textbf{Manip} & \textbf{Refuse} & \textbf{Avg.} \\
    \midrule
    \rowcolor{gray!20} 
    \multicolumn{8}{l}{\textit{Proprietary Model}}\\
    Gemini-2.5-Pro~\cite{gemini} & - & 59.8 & 45.5 & 49.0 & 33.6 & 38.9 & 45.2 \\
    Operator~\cite{cua} & - & 51.3 & 42.4 & 46.6 & 31.5 & 0.0 & 40.6 \\
    \midrule
    \rowcolor{gray!20} 
    \multicolumn{8}{l}{\textit{Fine-Tuning based GUI Agents}}\\
    Qwen2.5-VL-3B~\cite{qwen2.5} & - & 41.4 & 28.8 & 34.8 & 13.4 & 0.0 & 27.3 \\
    Qwen2.5-VL-7B~\cite{qwen2.5} & - & 45.6 & 32.7 & 41.9 & 18.1 & 0.0 & 31.4 \\
    UI-TARS-7B~\cite{uitars} & - & 60.2 & 51.8 & 54.9 & 35.6 & 0.0 & 47.5 \\
    UGround-7B~\cite{uground} & 10M & 51.3 & 40.3 & 43.5 & 24.8 & 0.0 & 36.4 \\
    Aguvis-7B~\cite{aguvis} & 1.036M & 55.9 & 41.2 & 43.9 & 28.2 & 0.0 & 38.7 \\
    GUI-Actor-7B~\cite{guiactor} & 1.2M & 60.2 & 54.2 & \underline{58.1} & 30.9 & 0.0 & 49.5 \\
    Jedi-3B~\cite{jedi} & 4M & \textbf{67.4} & 53.0 & 53.8 & \underline{44.3} & 7.4 & 50.9 \\
    Jedi-7B~\cite{jedi} & 4M & \underline{65.9} & \underline{55.5} & 57.7 & \textbf{46.9} & 7.4 & 54.1 \\
    FocusUI-7B~\cite{focusui} & 1M & 63.6 & \textbf{61.2} & \textbf{63.6} & 34.9 & 0.0 & \textbf{54.4} \\
    \midrule
     \rowcolor[HTML]{E8F0F8} \textbf{\name-3B}  & \textbf{0} & 53.3 & 46.4 & 44.3 & 36.4 & 0.4 & 49.9 \\
     \rowcolor[HTML]{E8F0F8} \textbf{\name-7B}  & \textbf{0} & 57.9 & 55.2 & 55.7 & 43.0 & 1.1 & \underline{54.2}  \\
    \bottomrule
    \end{tabular}
}
\label{osworld}
\vspace{-2ex}
\end{table}

Among these benchmarks, our proposed \name achieves competitive performance on ScreenSpot-Pro, OSWorld-G, and UI-Vision. However, its performance on MMBench-GUI is relatively insufficient. We attribute this variance to the inherent characteristics of the datasets. The former three primarily feature high-resolution web interfaces (e.g., 2K, 4K, or even 6K) paired with relatively simple and ambiguous instructions. These conditions naturally favor \name, which is explicitly designed to refine instructions and progressively zoom into target element regions. Conversely, MMBench-GUI provides explicit instructions and moderate resolutions (e.g., 1080p to 2K). By circumventing drawbacks mentioned before, this benchmark falls within their training distribution, hence readily exploiting their parameterized UI knowledge to familiar visual elements.

\begin{table}[t!]
\centering
\caption{Performance comparison on the \textbf{UI-Vision} benchmark. \textbf{Bold} and
\underline{underline} represents the best and second results, respectively.}
\resizebox{0.98\textwidth}{!}{
    \setlength{\tabcolsep}{2pt}
    \begin{tabular}{l | c | cccccc | ccc | c}
    \toprule
    \multirow{2}{*}{\textbf{Model}} & \multirow{2}{*}{\makecell{\textbf{Data}\\\textbf{Size}}} & \multicolumn{6}{c}{\textbf{Grouped by Category}} & \multicolumn{3}{c}{\textbf{Grouped by Setting}} & \multirow{2}{*}{\textbf{Avg.}} \\
    \cmidrule(lr){3-8} \cmidrule(lr){9-11}
     & & Edu. & Browser & Dev. & Prod. & Creative & Entert. & Basic & Func. & Spatial & \\
    \midrule
    \rowcolor{gray!20} 
    \multicolumn{12}{l}{\textit{Proprietary Model}}\\
    GPT-4o~\cite{gpt4.1} & - & 1.5 & 0.0 & 2.2 & 1.1 & 0.8 & 4.2 & 1.6 & 1.5 & 1.0 & 1.4 \\
    Claude~\cite{claude} & - & 6.1 & 9.8 & 8.0 & 9.4 & 7.7 & 8.3 & 9.5 & 7.7 & 7.6 & 8.3 \\
    \midrule
    \rowcolor{gray!20} 
    \multicolumn{12}{l}{\textit{Fine-Tuning based GUI Agents}}\\
    Qwen-2.5VL-7B~\cite{qwen2.5} & - & 0.5 & 0.0 & 1.2 & 0.9 & 0.5 & 1.0 & 1.2 & 0.8 & 0.5 & 0.9 \\
    InternVL2.5-8B~\cite{internvl} & - & 1.1 & 7.0 & 3.0 & 1.8 & 1.2 & 5.2 & 2.5 & 2.8 & 1.0 & 2.1 \\
    MiniCPM-V-8B~\cite{minicpm} & - & 3.0 & 16.8 & 5.4 & 3.8 & 2.1 & 13.0 & 7.1 & 5.3 & 1.5 & 4.3 \\
    SeeClick-9.6B~\cite{seeclick} & 1M & 4.2 & 13.3 & 7.3 & 4.3 & 4.0 & 11.0 & 9.4 & 4.7 & 2.1 & 5.4 \\
    ShowUI-2B~\cite{showui} & 256K & 3.7 & 13.3 & 7.5 & 6.5 & 2.5 & 15.6 & 8.1 & 7.7 & 2.1 & 5.9 \\
    CogAgent-9B~\cite{cogagent} & 400K & 8.7 & 11.2 & 8.6 & 10.3 & 5.6 & 15.6 & 12.0 & 12.2 & 2.6 & 8.9 \\
    OSAtlas-7B~\cite{osatlas} & 2.24M & 8.7 & 16.8 & 10.3 & 9.2 & 5.6 & 16.2 & 12.2 & 11.2 & 3.7 & 9.0 \\
    AriaUI-3B~\cite{ariaui} & 11.5M & 9.0 & 18.9 & 11.2 & 10.4 & 6.5 & 19.3 & 12.2 & 14.0 & 4.0 & 10.1 \\
    UGround-7B~\cite{uground} & 10M & 10.4 & 28.7 & 17.5 & 12.2 & 8.6 & 18.2 & 15.4 & 17.1 & 6.3 & 12.9 \\
    Aguvis-7B~\cite{aguvis} & 1.036M & 13.1 & 30.8 & 17.1 & 12.1 & 9.6 & 24.0 & 17.8 & 18.3 & 5.1 & 13.7 \\
    UI-TARS-7B~\cite{uitars} & - & 14.2 & 35.0 & 19.7 & 18.3 & 11.1 & 38.5 & 20.1 & 24.3 & 8.4 & 17.6 \\
    GUI-Actor-3B~\cite{guiactor} & 1.2M & - & - & - & - & - & - & 27.4 & 24.6 & 7.0 & 19.3 \\
    GUI-Actor-7B~\cite{guiactor} & 1.2M & - & - & - & - & - & - & 30.1 & 28.1 & 7.8 & 21.6 \\
    Jedi-3B~\cite{jedi} & 4M & - & - & - & - & - & - & 22.3 & 25.2 & 9.4 & 18.7 \\
    Jedi-7B~\cite{jedi} & 4M & - & - & - & - & - & - & 32.3 & \underline{30.5} & \underline{12.8} & 24.8 \\
    FocusUI-3B~\cite{focusui} & 1M & - & - & - & - & - & - & 30.0 & 26.9 & 8.7 & 21.5 \\
    FocusUI-7B~\cite{focusui} & 1M & - & - & - & - & - & - & 32.3 & 29.2 & 11.0 & 24.9 \\
    InfiGUI-G1-3B~\cite{infigui} & 32K & 22.6 & 43.4 & 24.3 & 22.6 & 14.0 & 47.4 & 31.2 & 28.0 & 8.2 & 22.0 \\
    InfiGUI-G1-7B~\cite{infigui} & 32K & \underline{25.5} & \textbf{46.2} & \underline{29.6} & \underline{26.7} & \underline{17.6} & \textbf{52.1} & \textbf{36.2} & \textbf{31.9} & 11.5 & \underline{26.1} \\
    \midrule
     \rowcolor[HTML]{E8F0F8} \textbf{\name-3B}  & \textbf{0} & 23.0 & 33.1 & 18.9 & 19.0 & 13.2 & 40.0 & 24.5 & 19.1 & 6.7 & 18.5 \\
     \rowcolor[HTML]{E8F0F8} \textbf{\name-7B}  & \textbf{0} & \textbf{33.2} & \underline{43.7} & \textbf{31.4} & \textbf{27.9} & \textbf{19.0} & \underline{49.3} & \underline{33.3} & 29.1 & \textbf{14.0} & \textbf{27.1} \\
    \bottomrule
    \end{tabular}
}
\label{uivision}
\vspace{-2ex}
\end{table}

\begin{table}[t!]
\centering
\caption{Performance comparison on \textbf{MMBench-GUI} benchmarks. \textbf{Bold} and \underline{underline} represents the best and second results, respectively.}
\resizebox{0.98\textwidth}{!}{
    \setlength{\tabcolsep}{2.5pt} 
    \begin{tabular}{l | c | cc cc cc cc cc cc | c}
    \toprule
    \multirow{2}{*}{\textbf{Model}} & \multirow{2}{*}{\makecell{\textbf{Data}\\\textbf{Size}}} & \multicolumn{2}{c}{\textbf{Windows}} & \multicolumn{2}{c}{\textbf{MacOS}} & \multicolumn{2}{c}{\textbf{Linux}} & \multicolumn{2}{c}{\textbf{iOS}} & \multicolumn{2}{c}{\textbf{Android}} & \multicolumn{2}{c|}{\textbf{Web}} & \multirow{2}{*}{\textbf{Avg.}} \\
    \cmidrule(lr){3-4} \cmidrule(lr){5-6} \cmidrule(lr){7-8} \cmidrule(lr){9-10} \cmidrule(lr){11-12} \cmidrule(lr){13-14}
     & & Bas. & Adv. & Bas. & Adv. & Bas. & Adv. & Bas. & Adv. & Bas. & Adv. & Bas. & Adv. & \\
    \midrule
    \rowcolor{gray!20} 
    \multicolumn{15}{l}{\textit{Proprietary Model}}\\
    GPT-4o~\cite{gpt4.1} & - & 1.5 & 1.1 & 8.7 & 4.3 & 1.1 & 1.0 & 5.1 & 3.3 & 2.5 & 1.4 & 3.2 & 2.9 & 2.9 \\
    Claude~\cite{claude} & - & 1.5 & 0.7 & 12.5 & 7.5 & 1.1 & 0.0 & 13.7 & 10.6 & 1.4 & 1.4 & 3.2 & 2.3 & 4.7 \\
    \midrule
    \rowcolor{gray!20} 
    \multicolumn{15}{l}{\textit{Fine-Tuning based GUI Agents}}\\
    ShowUI-2B~\cite{showui} & 256K & 9.2 & 4.4 & 24.1 & 10.4 & 25.1 & 11.7 & 29.0 & 19.7 & 17.4 & 8.7 & 22.9 & 12.7 & 16.0 \\
    Qwen2.5-VL-7B~\cite{qwen2.5} & - & 31.4 & 16.5 & 31.3 & 22.0 & 21.5 & 12.2 & 66.6 & 55.2 & 35.1 & 35.2 & 40.3 & 32.5 & 33.9 \\
    OSAtlas-7B~\cite{osatlas} & 2.24M & 36.9 & 18.8 & 44.4 & 21.7 & 31.4 & 13.3 & 74.8 & 48.8 & 69.6 & 46.8 & 61.3 & 35.4 & 41.4 \\
    Aguvis-7B~\cite{aguvis} & 1.036M & 37.3 & 21.7 & 48.1 & 33.3 & 33.5 & 25.0 & 67.5 & 65.2 & 61.0 & 51.0 & 61.6 & 45.5 & 45.7 \\
    UI-TARS-1.5-7B~\cite{uitars} & - & 68.3 & 39.0 & 69.0 & 44.5 & 64.4 & 37.8 & 88.5 & 69.4 & 90.5 & 69.3 & 81.0 & 56.5 & 64.3 \\
    UGround-7B~\cite{uground} & 10M & 66.8 & 39.0 & 71.3 & 48.6 & 56.5 & 31.1 & 92.7 & 70.9 & 93.5 & 71.0 & 88.7 & 64.6 & 65.7 \\
    GUI-Actor-7B~\cite{guiactor} & 1.2M & \underline{80.8} & 55.1 & 81.4 & 60.4 & 64.9 & 41.8 & 94.3 & 82.7 & 93.5 & 79.7 & 89.7 & 72.1 & 76.5 \\
    SE-GUI-7B~\cite{segui} & 3K & 77.5 & 57.7 & 77.1 & 60.7 & 68.6 & 44.9 & \textbf{95.5} & 80.0 & 95.5 & 83.7 & 89.7 & 68.8 & 76.6 \\
    GTA1-7B~\cite{gta1} & 1.56M & 76.8 & 57.4 & 80.3 & 63.9 & 68.6 & \textbf{53.6} & 93.9 & 83.3 & \underline{96.3} & 84.5 & 90.3 & 74.7 & 78.5 \\
    GUI-G$^2$-7B~\cite{guig2} & 100K & 79.7 & 55.1 & 79.7 & \underline{64.7} & 69.6 & 50.0 & \underline{95.2} & 82.7 & \textbf{96.6} & 85.4 & 91.9 & 75.6 & 78.8 \\
    InfiGUI-G1-7B~\cite{infigui} & 32K & \textbf{82.7} & \underline{61.8} & \underline{83.8} & 63.9 & 72.3 & \underline{52.0} & 94.9 & \underline{89.4} & 95.2 & \underline{85.6} & \textbf{93.5} & \underline{76.3} & \underline{80.8} \\
    UI-Ins-7B~\cite{ui-ins} & 316K & \textbf{82.7} & \textbf{64.7} & \textbf{87.2} & \textbf{75.1} & 71.7 & 51.5 & 94.9 & \textbf{89.7} & 95.8 & \textbf{89.0} & \underline{93.2} & \textbf{80.8} & \textbf{83.1} \\
    \midrule
    \rowcolor[HTML]{E8F0F8} \textbf{\name-3B} & \textbf{0} & 65.1 & 43.1 & 71.4 & 48.0 & \underline{75.9} & 38.3 & 77.9 & 70.4 & 76.9 & 70.5 & 65.0 & 62.9 & 63.2 \\
    \rowcolor[HTML]{E8F0F8} \textbf{\name-7B} & \textbf{0} & 73.6 & 49.6 & 77.6 & 60.8 & \textbf{78.5} & 47.7 & 75.3 & 78.2 & 76.6 & 79.9 & 75.4 & 69.9 & 72.8 \\
    \bottomrule
    \end{tabular}
    \label{mmbench}
}
\vspace{-2ex}
\end{table}

\subsection{Ablation Studies}
\subsubsection{Effect of Instruction Refinement.} We conduct the ablation study with two experiments, the first is fully discard instruction refining and directly leverage original instruction for GUI grounding (w/o Refinement), and the second is to directly leverage MLLM's capacity to generate refined instructions without thinking process (w/o Think). We select ScreenSpot-Pro and UI-Vision to evaluate and results are shown in Table~\ref{irablation}. We observe that w/o Refinement leads to a significant performance degradation, particularly for the 3B MLLM. We argue that smaller-parameter models are more susceptible to ambiguous instructions, which impair their relatively weaker GUI grounding capability. Furthermore, while leveraging MLLM's ability to refine instruction yields performance improvements, it is still worse than integrating with thinking process, which verifies the effectiveness of our proposed internal thinking mechanism.

\begin{table}[t!]
\centering
\caption{Instruction refinement ablation study on the \textbf{ScreenSpot-Pro} and \textbf{UI-Vision} benchmark. The \textcolor{gray}{gray} are main results for reference.}
\resizebox{0.95\textwidth}{!}{
    \setlength{\tabcolsep}{3pt}
    \begin{tabular}{l | cc cc cc cc cc cc | c}
    \toprule

    \multirow{2}{*}{\textbf{ScreenSpot-Pro}} & \multicolumn{2}{c}{\textbf{CAD}} & \multicolumn{2}{c}{\textbf{Dev.}} & \multicolumn{2}{c}{\textbf{Creative}} & \multicolumn{2}{c}{\textbf{Scientific}} & \multicolumn{2}{c}{\textbf{Office}} & \multicolumn{2}{c|}{\textbf{OS}} & \multirow{2}{*}{\textbf{Avg.}} \\
    \cmidrule(lr){2-3} \cmidrule(lr){4-5} \cmidrule(lr){6-7} \cmidrule(lr){8-9} \cmidrule(lr){10-11} \cmidrule(lr){12-13}
     & Text & Icon & Text & Icon & Text & Icon & Text & Icon & Text & Icon & Text & Icon & \\
    \midrule
    \rowcolor[HTML]{F2F2F2} \textbf{\name-3B} & 47.72 & 20.3 & 52.6 & 18.6 & 45.5 & 14.0 & 52.8 & 25.5 & 62.2 & 37.7 & 54.2 & 22.5 & 40.2 \\
    \midrule
    \rowcolor[HTML]{F2F2F2} \textbf{\name-7B} & 59.9 & 25.0 & 64.9 & 30.3 & 56.6 & 23.1 & 67.4 & 31.8 & 81.4 & 43.4 & 67.3 & 34.8 & 52.2 \\
    \midrule
    \textbf{\name-3B} \\ w/o Refinement & 20.8 & 3.1 & 16.9 & 5.5 & 18.2 & 5.6 & 29.9 & 5.5 & 24.9 & 3.8 & 18.7 & 10.1 & 15.5 \\
    \textbf{\name-7B} \\ w/o Refinement & 54.3 & 25.0 & 48.7 & 33.8 & 45.5 & 27.3 & 52.1 & 31.8 & 68.9 & 43.4 & 49.5 & 29.2 & 44.9 \\
    \midrule
    \textbf{\name-3B} \\ w/o Think & 47.2 & 20.3 & 50.7 & 20.7 & 48.5 & 13.3 & 56.3 & 16.4 & 64.9 & 35.9 & 46.7 & 21.4 & 39.9 \\
    \textbf{\name-7B} \\ w/o Think & 59.3 & 23.4 & 53.2 & 26.9 & 48.9 & 24.5 & 55.6 & 32.7 & 67.2 & 43.4 & 57.0 & 34.8 & 46.5 \\

    \midrule[\heavyrulewidth]
    \multirow{2}{*}{\textbf{UI-Vision}} & 
    \multicolumn{6}{c}{\textbf{Grouped by Category}} & \multicolumn{3}{c}{\textbf{-}} & \multicolumn{3}{c|}{\textbf{Grouped by Setting}} & \multirow{2}{*}{\textbf{Avg.}} \\
    \cmidrule(lr){2-7} \cmidrule(lr){11-13} 
    
     & Edu. & Brow. & Dev. & Prod. & Crea. & Ent. & - & - & - & Basic & Func. & Spat. & \\
    \midrule
    
    \rowcolor[HTML]{F2F2F2} \textbf{UI-Infer-3B} & 23.0 & 33.1 & 18.9 & 19.0 & 13.2 & 40.0  & - & - & - & 24.5 & 19.1 & 6.7 & 18.5 \\
    \rowcolor[HTML]{F2F2F2} \textbf{UI-Infer-7B} & 33.2 & 43.7 & 31.4 & 27.9 & 19.0 & 49.3 & - & - & - & 33.3 & 29.1 & 14.0 & 27.1 \\
    \midrule
    \textbf{UI-Infer-3B} \\ w/o Refinement & 25.2 & 27.2 & 18.0 & 17.6 & 12.7 & 46.7 & - & - & - & 23.5 & 17.7 & 7.4 & 17.4 \\
    \textbf{UI-Infer-7B} \\ w/o Refinement & 27.2 & 33.9 & 21.3 & 21.4 & 10.7 & 41.3 & - & - & - & 25.9 & 22.0 & 10.0 & 21.4 \\
    \midrule
    \textbf{UI-Infer-3B} \\ w/o Think & 24.3 & 29.5 & 17.0 & 19.3 & 12.5 & 36.0 & - & - & - & 22.9 & 17.9 & 7.8 & 17.7 \\
    \textbf{UI-Infer-7B} \\ w/o Think & 31.9 & 41.8 & 29.4 & 25.0 & 19.5 & 53.3 & - & - & - & 31.6 & 28.9 & 11.5 & 25.3 \\
    \bottomrule
    \end{tabular}
}
\label{irablation}
\vspace{-2ex}
\end{table}

\subsubsection{Effect of Visual Focus.}
We then validate the effect of visual focus (w/o VF), and the results are shown in Table~\ref{vfablation}. As observed, w/o VF severely degrades GUI grounding performance across both the 3B and 7B MLLMs. This demonstrates that the dense information and high resolutions of UI interfaces substantially impede models' ability to locate target elements. Furthermore, this performance drop is notably more drastic than that observed in the w/o Refinement setting (Table~\ref{irablation}), leading us to conclude that fine-grained visual comprehension plays a more dominant role in GUI grounding tasks.

\begin{table}[t!]
\centering
\caption{Visual focus ablation study on the \textbf{ScreenSpot-Pro} and \textbf{UI-Vision} benchmark. The \textcolor{gray}{gray} are main results for reference.}
\resizebox{0.95\textwidth}{!}{
    \setlength{\tabcolsep}{3pt}
    \begin{tabular}{l | cc cc cc cc cc cc | c}
    \toprule

    \multirow{2}{*}{\textbf{ScreenSpot-Pro}} & \multicolumn{2}{c}{\textbf{CAD}} & \multicolumn{2}{c}{\textbf{Dev.}} & \multicolumn{2}{c}{\textbf{Creative}} & \multicolumn{2}{c}{\textbf{Scientific}} & \multicolumn{2}{c}{\textbf{Office}} & \multicolumn{2}{c|}{\textbf{OS}} & \multirow{2}{*}{\textbf{Avg.}} \\
    \cmidrule(lr){2-3} \cmidrule(lr){4-5} \cmidrule(lr){6-7} \cmidrule(lr){8-9} \cmidrule(lr){10-11} \cmidrule(lr){12-13}
     & Text & Icon & Text & Icon & Text & Icon & Text & Icon & Text & Icon & Text & Icon & \\
    \midrule
    \rowcolor[HTML]{F2F2F2} \textbf{\name-3B} & 47.72 & 20.3 & 52.6 & 18.6 & 45.5 & 14.0 & 52.8 & 25.5 & 62.2 & 37.7 & 54.2 & 22.5 & 40.2 \\
    \midrule
    \rowcolor[HTML]{F2F2F2} \textbf{\name-7B} & 59.9 & 25.0 & 64.9 & 30.3 & 56.6 & 23.1 & 67.4 & 31.8 & 81.4 & 43.4 & 67.3 & 34.8 & 52.2 \\
    \midrule
    \textbf{\name-3B} \\ w/o VF & 17.3 & 4.7 & 16.2 & 1.4 & 28.3 & 4.9 & 35.4 & 10.9 & 29.9 & 15.1 & 11.2 & 4.5 & 16.9 \\
    \textbf{\name-7B} \\ w/o VF & 15.2 & 3.1 & 35.1 & 4.1 & 21.2 & 5.6 & 37.5 & 5.5 & 43.5 & 13.2 & 28.0 & 6.7 & 20.4 \\

    \midrule[\heavyrulewidth]
    \multirow{2}{*}{\textbf{UI-Vision}} & 
    \multicolumn{6}{c}{\textbf{Grouped by Category}} & \multicolumn{3}{c}{\textbf{-}} & \multicolumn{3}{c|}{\textbf{Grouped by Setting}} & \multirow{2}{*}{\textbf{Avg.}} \\
    \cmidrule(lr){2-7} \cmidrule(lr){11-13}

     & Edu. & Brow. & Dev. & Prod. & Crea. & Ent. & - & - & - & Basic & Func. & Spat. & \\
    \midrule
    
    \rowcolor[HTML]{F2F2F2} \textbf{UI-Infer-3B} & 23.0 & 33.1 & 18.9 & 19.0 & 13.2 & 40.0  & - & - & - & 24.5 & 19.1 & 6.7 & 18.5 \\
    \rowcolor[HTML]{F2F2F2} \textbf{UI-Infer-7B} & 33.2 & 43.7 & 31.4 & 27.9 & 19.0 & 49.3 & - & - & - & 33.3 & 29.1 & 14.0 & 27.1 \\
    \midrule
    \textbf{UI-Infer-3B} \\ w/o VF & 15.5 & 26.6 & 14.6 & 11.7 & 7.8 & 33.3 & - & - & - & 14.8 & 12.0 & 3.2 & 12.5 \\
    \textbf{UI-Infer-7B} \\ w/o VF & 23.5 & 34.6 & 24.0 & 15.9 & 10.2 & 46.7 & - & - & - & 20.5 & 18.5 & 5.3 & 17.5\\
    \bottomrule
    \end{tabular}
}
\label{vfablation}
\vspace{-1ex}
\end{table}

\subsection{Discussion}

\subsubsection{Parameters Sensitivity.} 
\begin{figure*}[t!]
    \centering 
    
    \begin{subfigure}[b]{0.33\textwidth}
        \centering
        \includegraphics[width=\linewidth]{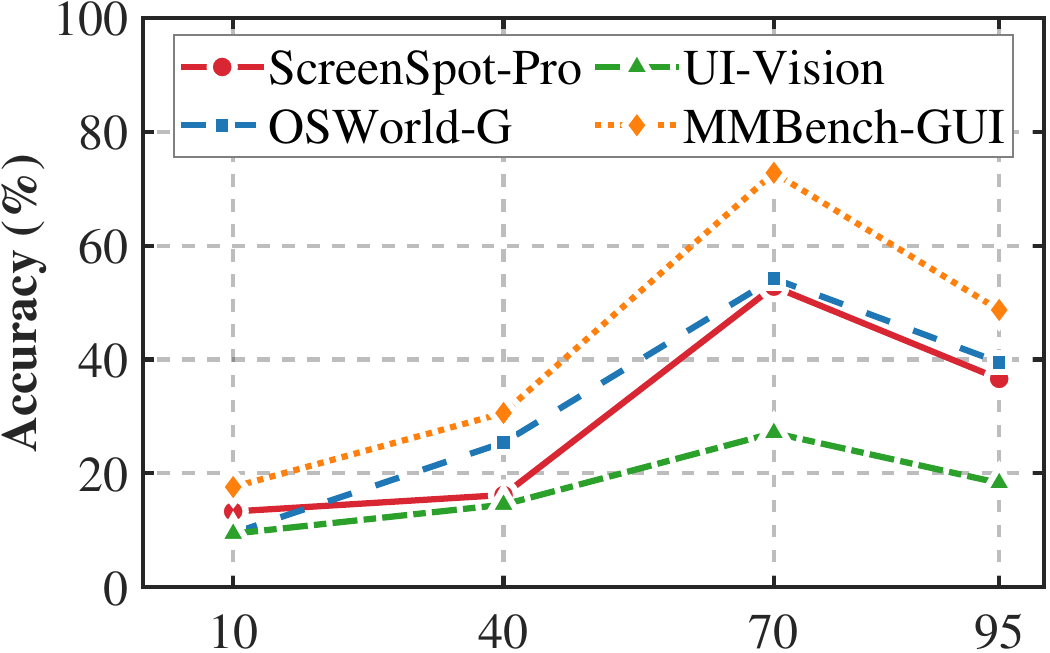}
        \caption{Proportion of layer depth (\%)}
        \label{fig:layer}
    \end{subfigure}%
\hfill %
    \begin{subfigure}[b]{0.33\textwidth}
        \centering
        \includegraphics[width=\linewidth]{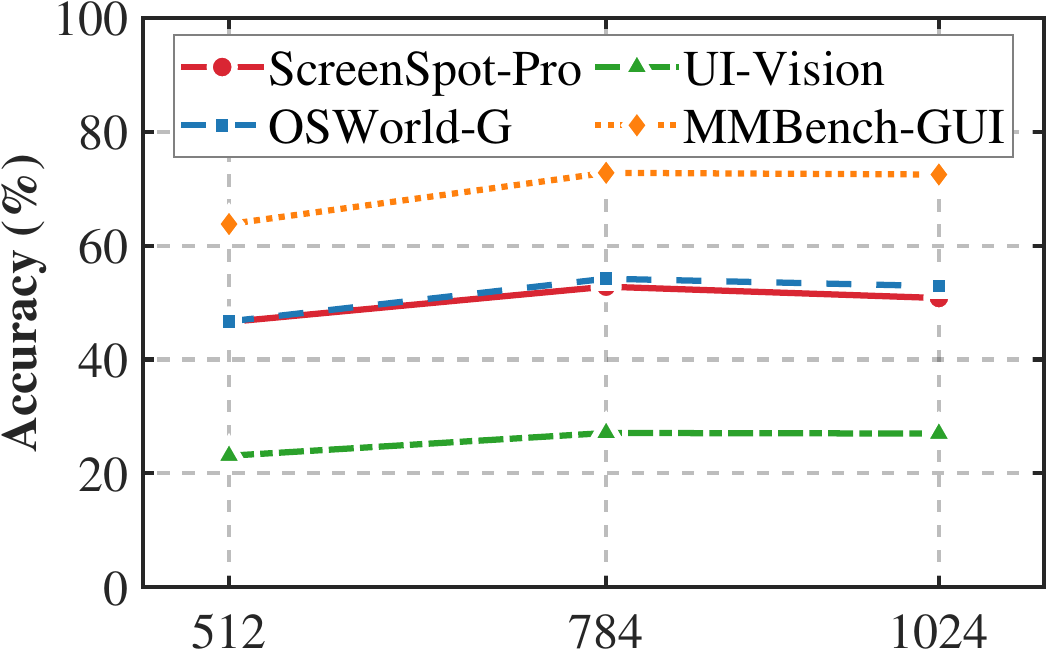}
        \caption{Window size}
        \label{fig:window}
    \end{subfigure}%
    \hfill %
    \begin{subfigure}[b]{0.33\textwidth}
        \centering
        \includegraphics[width=\linewidth]{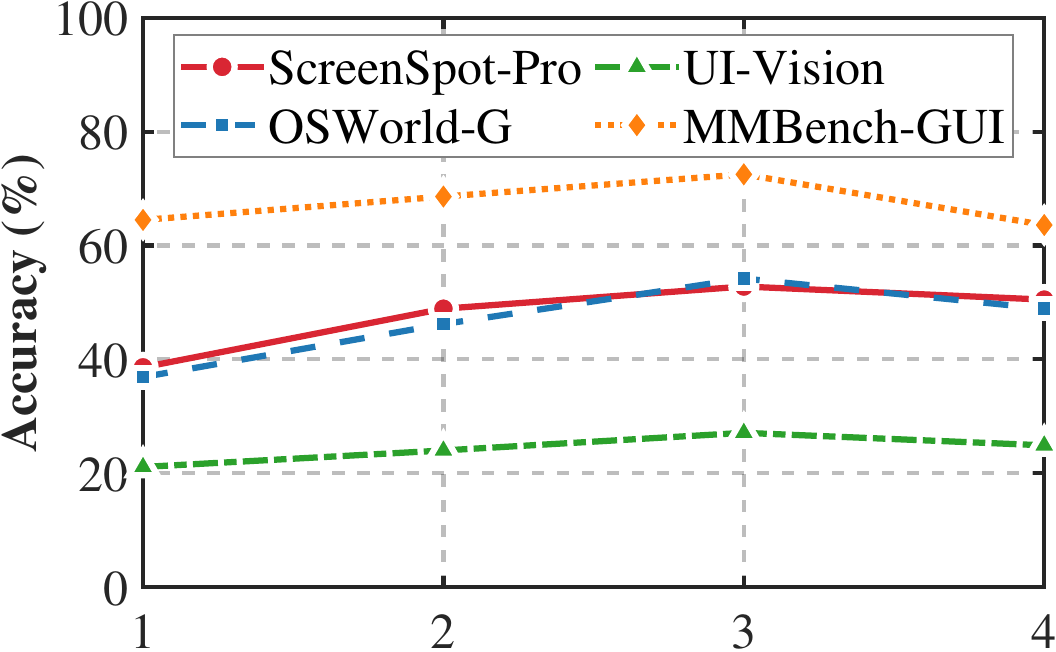}
        \caption{Iterations}
        \label{fig:iterations}
    \end{subfigure}%
    \hfill %

    \caption{Performance variance with different settings under four benchmarks.}
    
    \label{fig:sensitivity}
    \vspace{-2ex}
\end{figure*}

We first investigate the sensitivity of average grounding accuracy to layer depth chosen for capturing attention, we denote this depth as a percentage of the total layers, and results are illustrated in Figure~\ref{fig:layer}. Capturing attention from shallow or intermediate layer yields poor results, since the MLLM's understanding of target element remains incomplete. Conversely, final layers also degrade performance because they over-specialize for next-token prediction, shifting attention from visual spatial grounding to linguistic contexts. Secondly, as shown in Figure~\ref{fig:window}, a smaller window size (e.g., 512) does not guarantee better performance. We hypothesize that such a restricted visual field discards surrounding context, thereby impacting MLLM's ability to ground target element. Finally, in Figure~\ref{fig:iterations}, the results show that performance peaks at 3 iterations. This indicates that excessive zooming carries the risk of restricting essential visual context and cropping out the target element.

\begin{figure*}[t!]
    \centering 
    
    \begin{subfigure}[b]{0.33\textwidth}
        \centering
        \includegraphics[width=\linewidth]{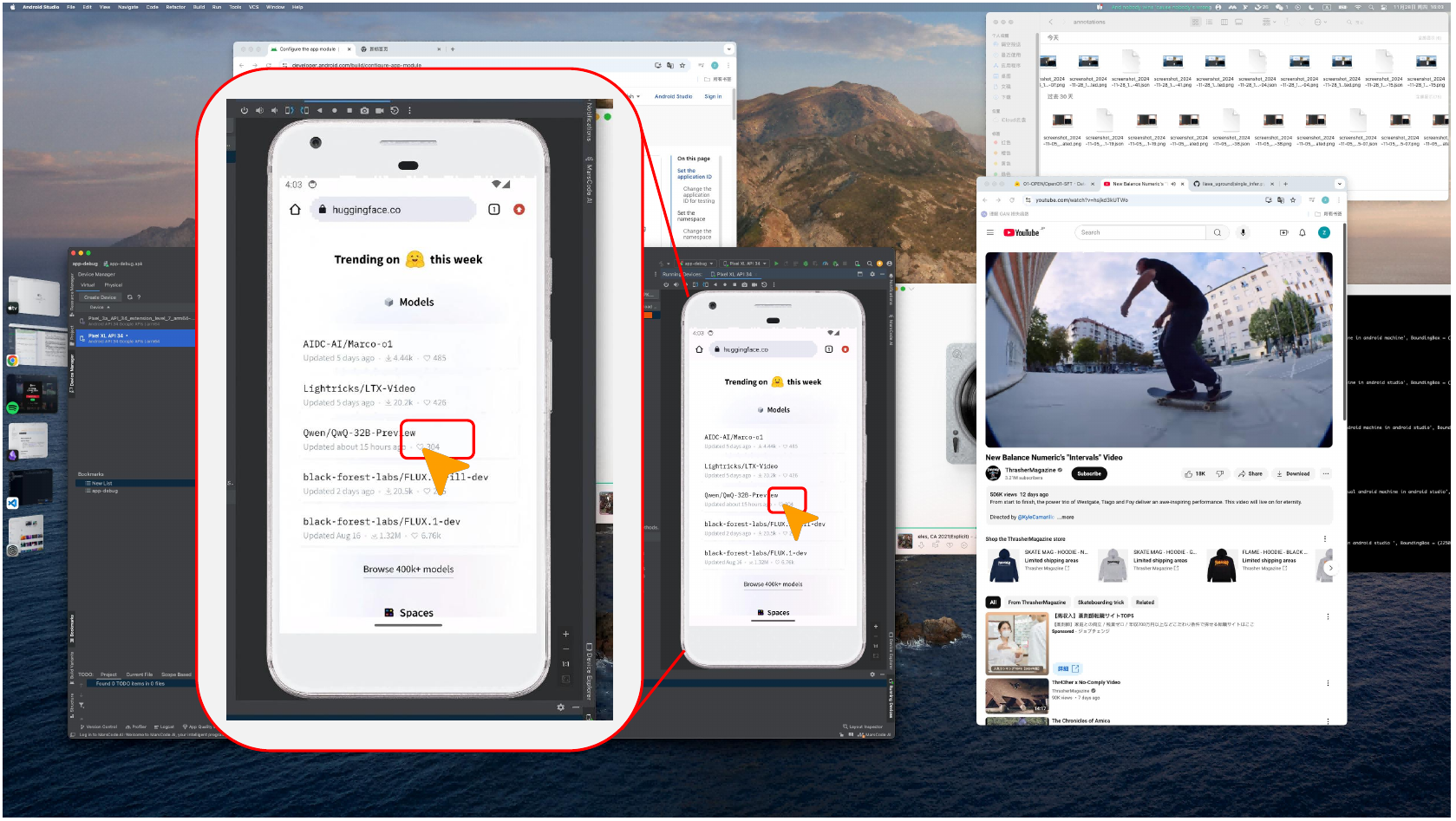}
        \caption{Raw UI interface}
        \label{fig:pro}
    \end{subfigure}%
\hfill %
    \begin{subfigure}[b]{0.33\textwidth}
        \centering
        \includegraphics[width=\linewidth]{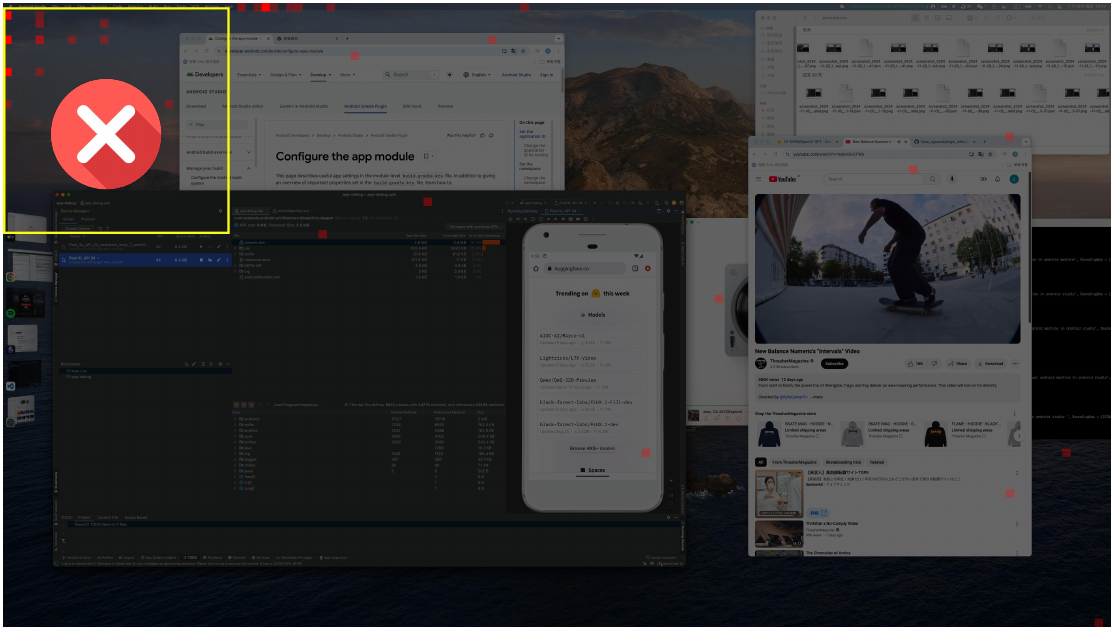}
        \caption{Prefill phase attention}
        \label{fig:prefill}
    \end{subfigure}%
    \hfill %
    \begin{subfigure}[b]{0.33\textwidth}
        \centering
        \includegraphics[width=\linewidth]{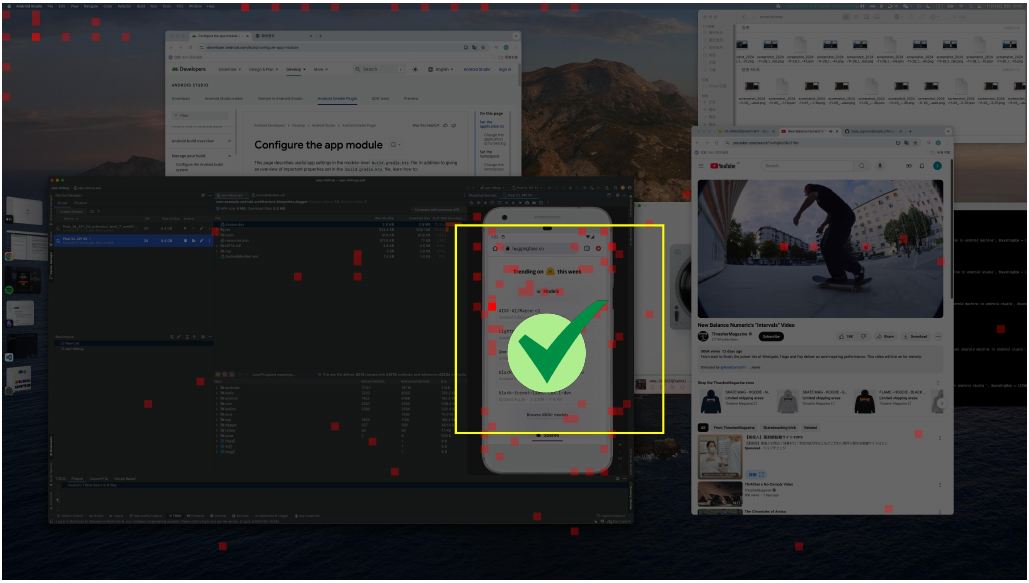}
        \caption{Generation phase attention}
        \label{fig:generation}
    \end{subfigure}%
    \hfill %

    \caption{Visualization comparison between prefill and generation phase of capturing attention distribution. (a) is UI interface with instruction: ``Like the QwQ-32b-preview on huggingface within the virtual android machine'', red block represents the ground truth region. Yellow blocks in attention maps of (b) and (c) are zoom-in regions refer to the highest attention scores.}
    
    \label{fig:captureatten}
    \vspace{-2ex}
\end{figure*}

\subsubsection{Phase of Capturing Attention.} \name captures MLLM's attention distribution during element coordinates generation (Generation). To validate the necessity, we investigate an alternative attempt: extracting attention during the cross-modal fusion stage 
In this phase, model focuses on aligning text instruction modality with visual interface modality, which primarily reflects MLLM's initial understanding of GUI grounding (Prefill), prior to making decision. We first show a visualization results in Figure~\ref{fig:captureatten} (we will show more visualization results in \textbf{Appendix B}). It can be observed that the generation phase attention accurately focuses on the region where target element exists, while prefill phase attention locate irrelevant area in the upper left corner.
\begin{wraptable}{l}{0.5\textwidth} 
\vspace{-28pt}
\caption{Performance comparison across different attention capture phases.}
\label{tab:attention_map_phases}
\setlength{\tabcolsep}{5pt}
\resizebox{0.45\textwidth}{!}{
    \begin{tabular}{l | cc}
    \toprule
    \textbf{Benchmark} & \textbf{Prefill} & \textbf{Generation} \\
    \midrule
    ScreenSpot-Pro & 18.1$\downarrow$ & 52.8 \\
    OSWorld-G      & 10.2$\downarrow$ & 54.2 \\
    UI-Vision      & 7.6$\downarrow$ & 27.1 \\
    MMBench-GUI & 17.3$\downarrow$ & 72.8 \\
    \bottomrule
    \end{tabular}
}
\label{attentionphase}
\vspace{-18pt}
\end{wraptable}
We then explore average accuracy results shown in Table~\ref{attentionphase}. It demonstrates that capturing the attention distribution during prefill phase for determining visual focus region severely degrades performance. We argue that MLLM only performs instruction and global interface alignment. Its generative intention is not formed, which means it lacks reasoning required to understand GUI elements. Consequently, cropping the region based on this unfocused attention discards actual target element, leading to failed grounding.

\subsubsection{Limitations.} The evaluations in main text are primarily based on the Qwen2.5-VL model. We will explore other advanced MLLMs in \textbf{Appendix C} to pursue further performance gains. Second, by freezing the MLLM's weights, our approach inevitably neglects advantages offered by data-driven fine-tuning. In future work, we aim to explore test-time adaptation strategies, selectively introducing UI data to dynamically adjust MLLM's parameters during inference phase, thereby further boosting the GUI grounding performance.

\section{Conclusion}
In this paper, we present \name, a training-free framework that achieves GUI grounding through inference scaling rather than costly data fine-tuning. By integrating latent thinking optimization for instruction refinement with an attention-guided visual focus mechanism, \name  handles the ambiguous user instructions and complex UI interface challenge in GUI grounding task. Extensive evaluations demonstrate that our approach rivals or surpasses SOTA fine-tuned baselines. Ultimately, \name highlights the immense potential of unlocking the inherent grounding capability of general MLLM for GUI automation.

\bibliographystyle{splncs04}
\bibliography{ref}

\clearpage

\begin{center}
  {\Large \bf Zoom to Essence: Trainless GUI Grounding by Inferring upon Interface Elements}\\[0.4cm]
  {\Large \bf Supplementary Material}
\end{center}
\vspace{1cm} 

\setcounter{section}{0}
\setcounter{figure}{0}
\setcounter{table}{0}
\setcounter{equation}{0}

\renewcommand{\thesection}{\Alph{section}.}

\section*{Abstract}

We provide additional content in supplementary material:\\

\textbf{(A)}~Discussing parameters sensitivity regarding Instruction Refinement. \\

\textbf{(B)}~Analyzing visualization results of attention capture phase in Visual Focus. \\

\textbf{(C)}~Exploring performance based on other advanced MLLMs. \\

\section{Parameters Sensitivity of Instruction Refinement}
\label{sec:parameters}

\begin{figure*}[htbp]
    \centering 
    
    \begin{subfigure}[b]{0.33\textwidth}
        \centering
        \includegraphics[width=\linewidth]{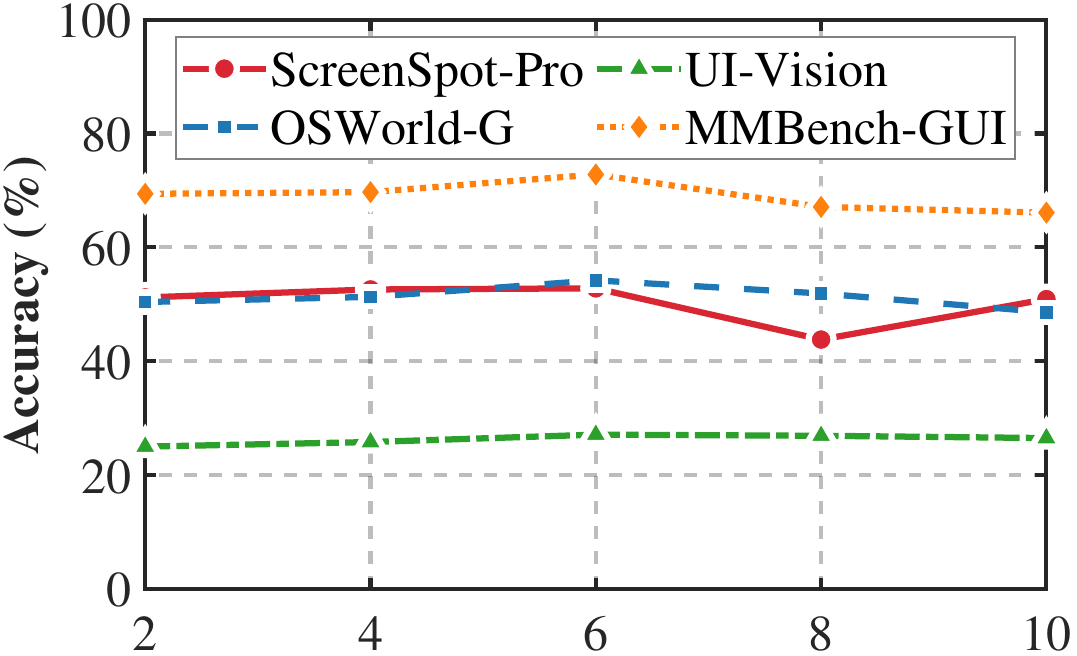}
        \caption{Length of thought vectors}
        \label{fig:vectors}
    \end{subfigure}%
\hfill %
    \begin{subfigure}[b]{0.33\textwidth}
        \centering
        \includegraphics[width=\linewidth]{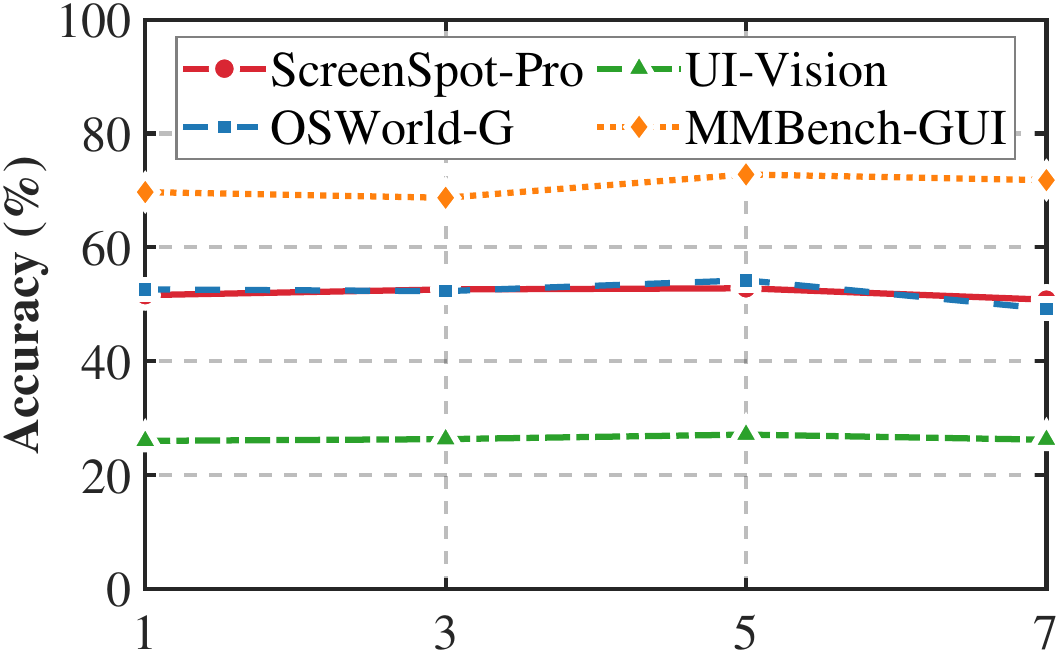}
        \caption{Thinking steps}
        \label{fig:steps}
    \end{subfigure}%
    \hfill %
    \begin{subfigure}[b]{0.33\textwidth}
        \centering
        \includegraphics[width=\linewidth]{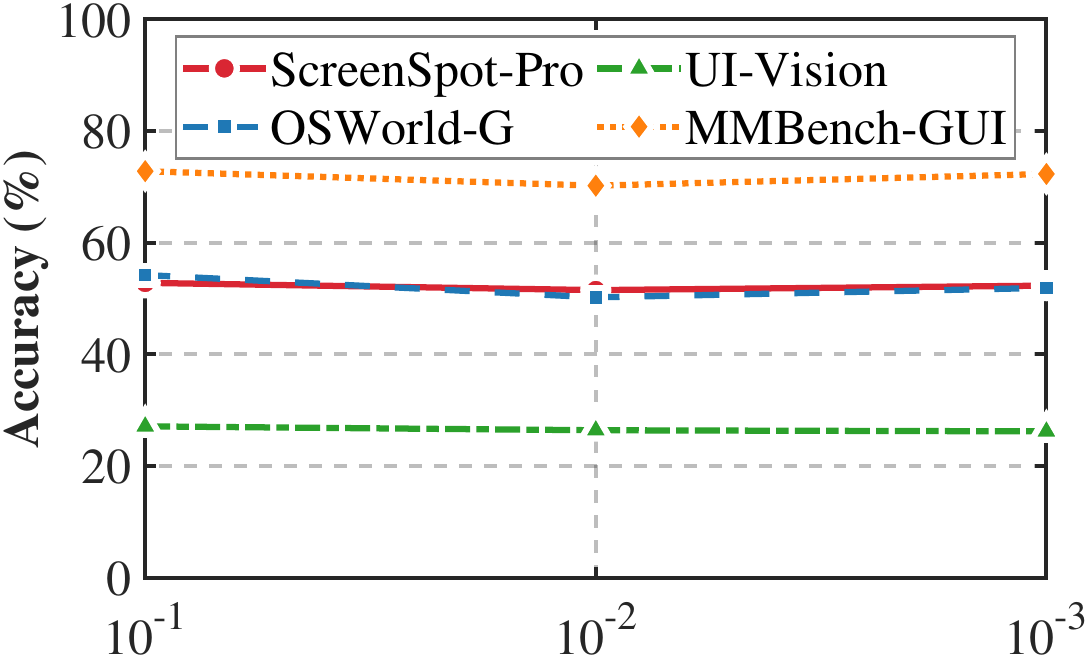}
        \caption{Learning rate}
        \label{fig:lr}
    \end{subfigure}%
    \hfill %

    \caption{Parameter sensitivity of instruction refinement across four benchmarks.}
    \label{fig:sensitivityappendix}
    \vspace{-2ex}
\end{figure*}

In this section, we provide a detailed analysis of this latent thinking strategy in Instruction Refinement: the sequence length of the injected learnable thought vectors (default $N=6$), the iterative thinking steps (default $K=5$), and the learning rate (default $\eta=10^{-1}$). We choose ZoomUI-7B to evaluate on multiple benchmarks, and results are shown in Figure~\ref{fig:sensitivityappendix}.

We can find the grounding accuracy across all four benchmarks reaches its peak when allocating an appropriate length of thought vectors ($N$) and depth of optimization steps ($K$). This optimal balance provides sufficient exploratory space for visual refinement while preventing noise from over-thinking. Interestingly, the performance exhibits minimal fluctuations regardless of variations in the learning rate ($\eta$). This insensitivity indicates that the gradient optimization has a straightforward convergence path, locating the correct visual features without relying on precisely tuned learning rates.

\begin{figure*}[t!]
    \centering 
    
    \begin{subfigure}[b]{0.325\textwidth}
        \centering
        \includegraphics[width=\linewidth]{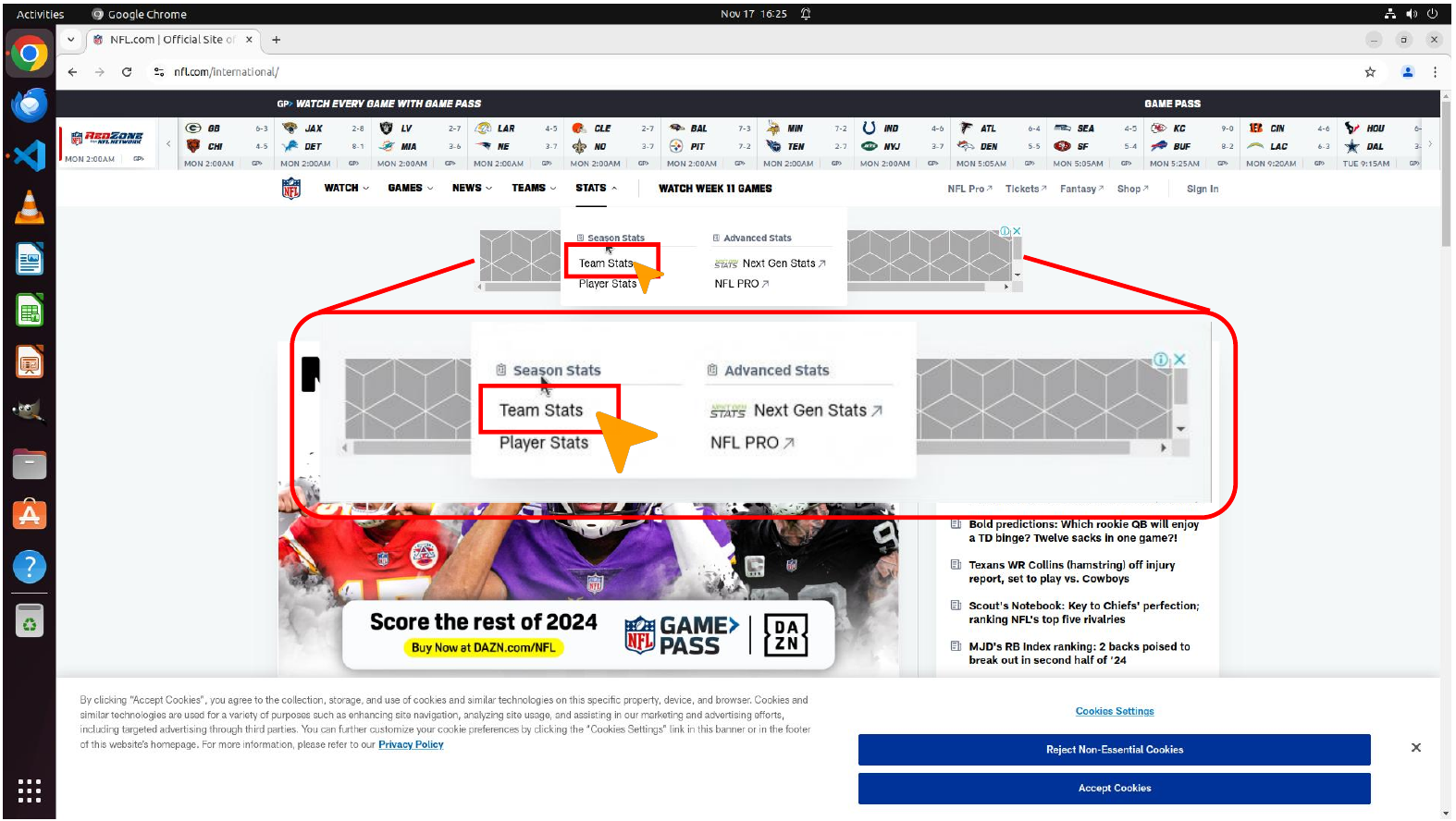}
        \caption{Raw Web interface}
        \label{fig:os1}
    \end{subfigure}%
\hfill %
    \begin{subfigure}[b]{0.325\textwidth}
        \centering
        \includegraphics[width=\linewidth]{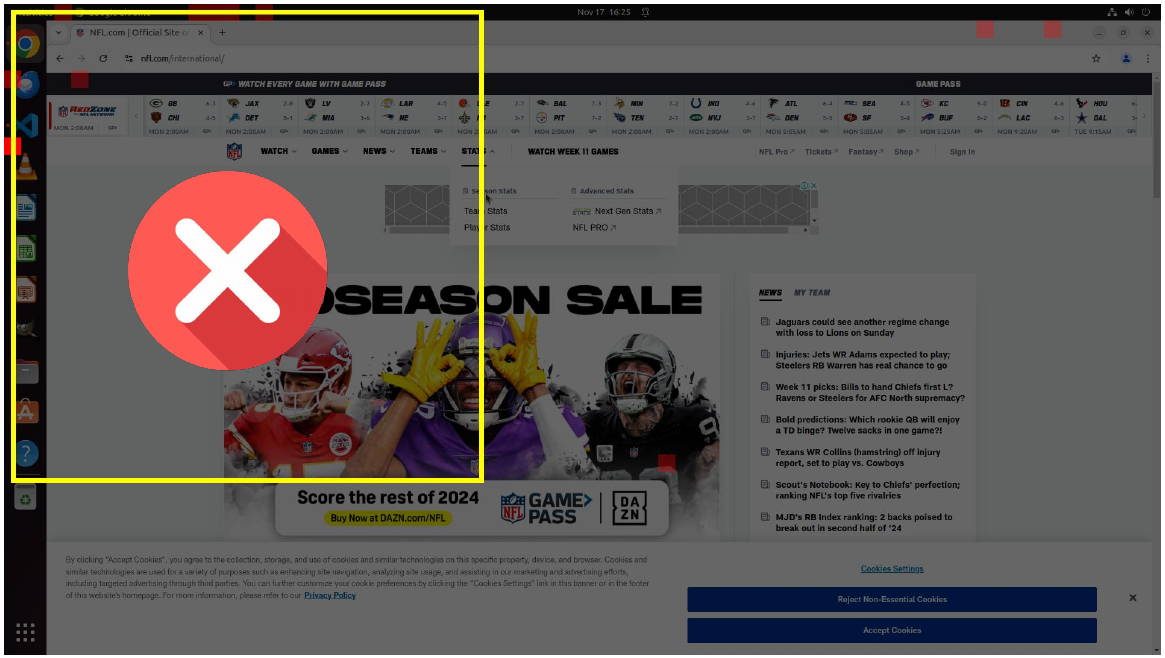}
        \caption{Prefill phase}
        \label{fig:osprefill}
    \end{subfigure}%
    \hfill %
    \begin{subfigure}[b]{0.325\textwidth}
        \centering
        \includegraphics[width=\linewidth]{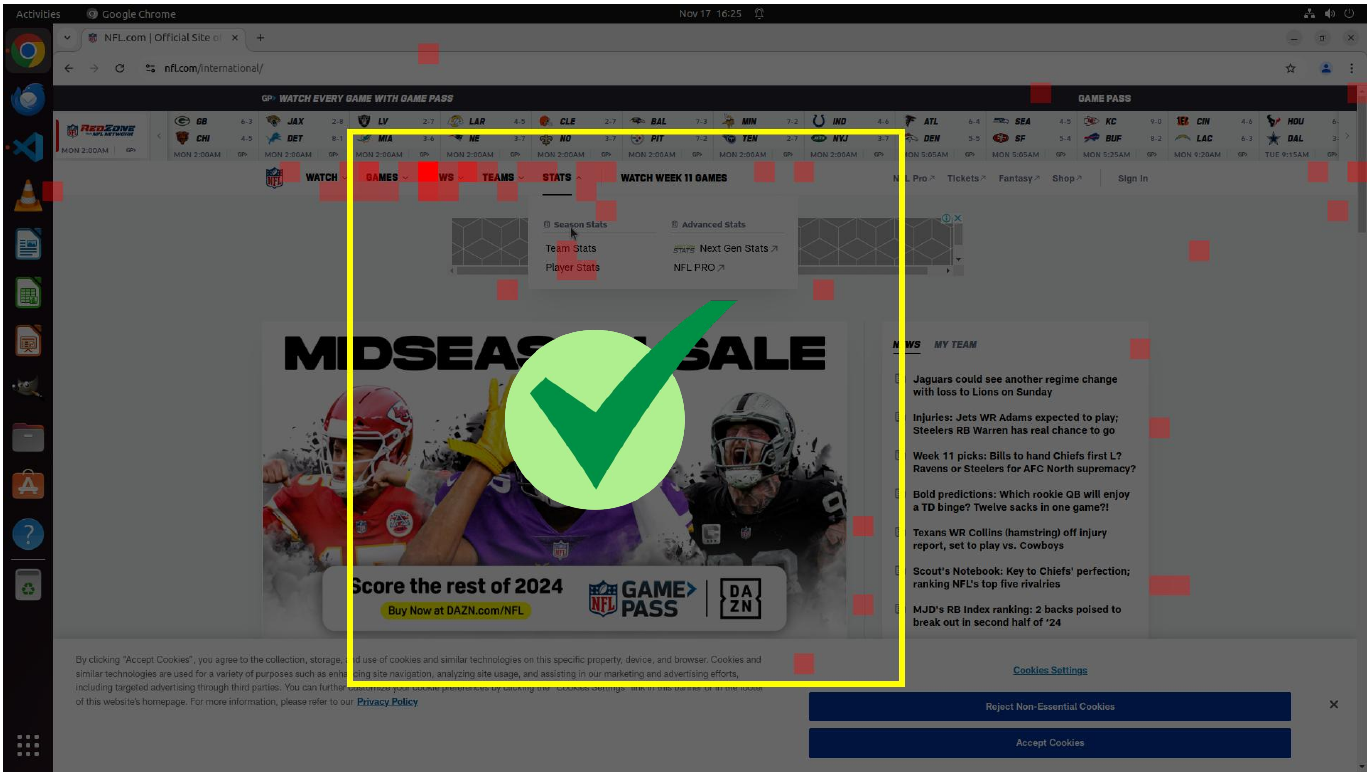}
        \caption{Generation phase}
        \label{fig:osgeneration}
    \end{subfigure}%
    \hfill %

    \caption{Visualization comparison. (a) is a UI interface with instruction: ``Check Stats of NFL teams'', red block represents the ground truth region. Yellow blocks in attention maps of (b) and (c) are zoom-in regions that refer to the highest attention scores.}
    
    \label{fig:oscaptureatten}
\end{figure*}

\begin{figure*}[t!]
    \centering 
    
    \begin{subfigure}[b]{0.25\textwidth}
        \centering
        \includegraphics[width=\linewidth]{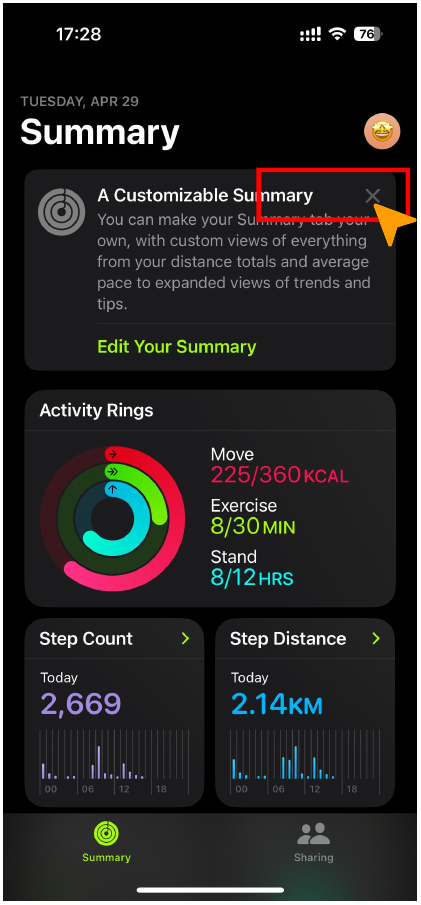}
        \caption{Raw iOS interface}
        \label{fig:ios}
    \end{subfigure}%
\hfill %
    \begin{subfigure}[b]{0.25\textwidth}
        \centering
        \includegraphics[width=\linewidth]{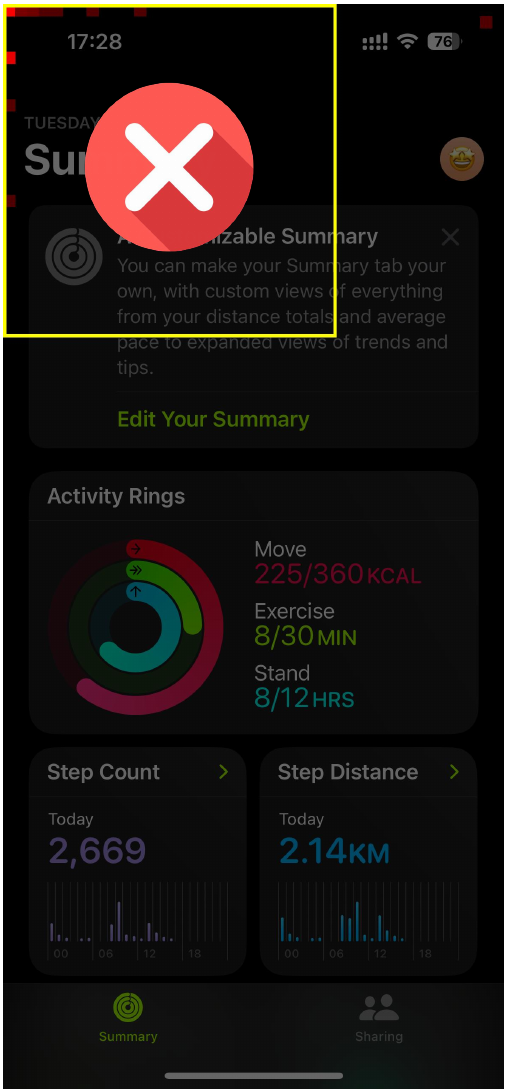}
        \caption{Prefill phase}
        \label{fig:iosprefill}
    \end{subfigure}%
    \hfill %
    \begin{subfigure}[b]{0.25\textwidth}
        \centering
        \includegraphics[width=\linewidth]{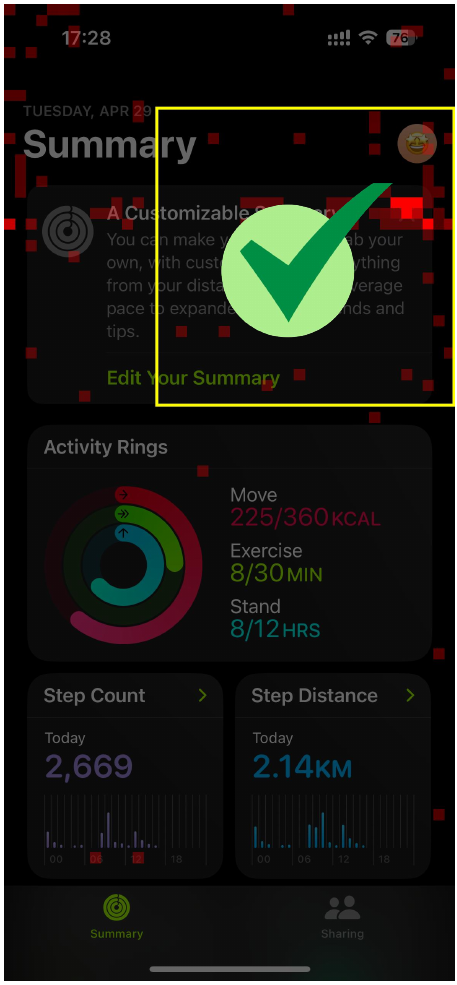}
        \caption{Generation phase}
        \label{fig:iosgeneration}
    \end{subfigure}%
    \hfill %

    \caption{Visualization comparison. (a) is a UI interface with instruction: ``Close the information prompt about personalizing your fitness dashboard'', red block represents the ground truth region. Yellow blocks in attention maps of (b) and (c) are zoom-in regions that refer to the highest attention scores.}
    \label{fig:ioscaptureatten}

\end{figure*}

\begin{figure*}[t!]
    \centering 
    
    \begin{subfigure}[b]{0.325\textwidth}
        \centering
        \includegraphics[width=\linewidth]{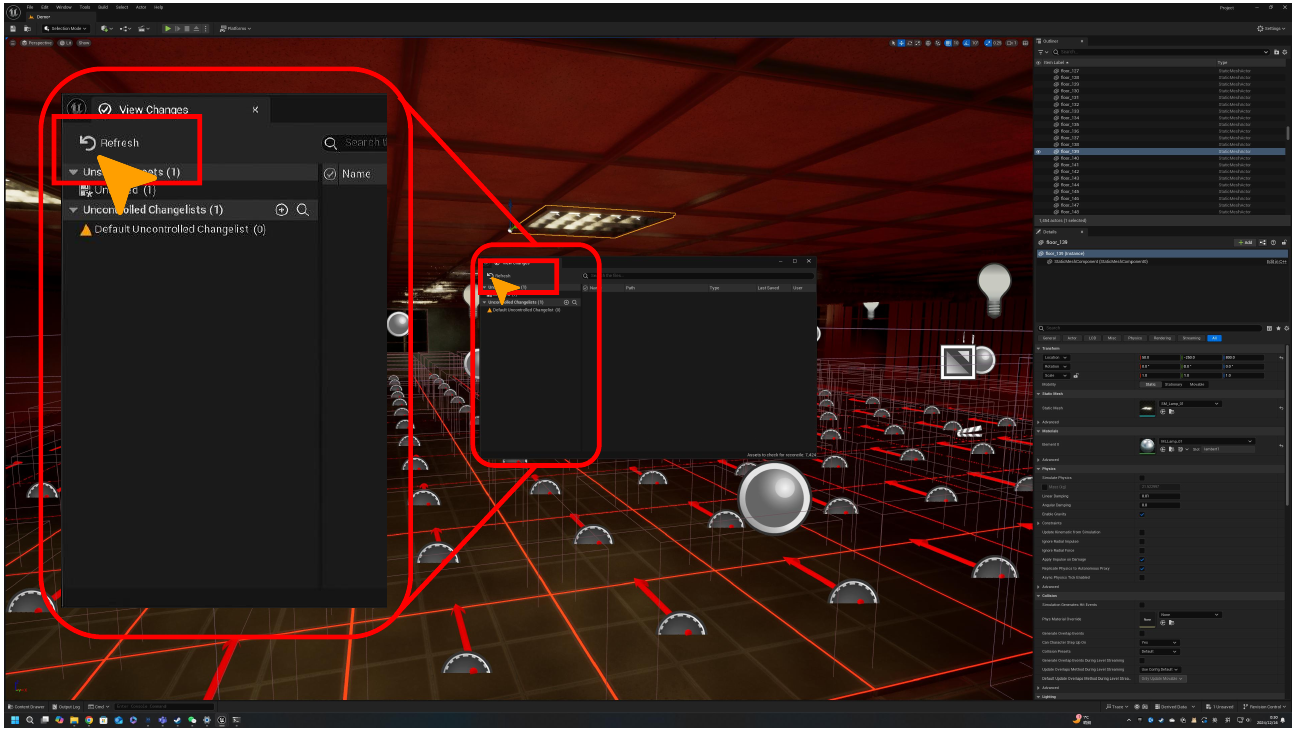}
        \caption{Raw software interface}
        \label{fig:un1}
    \end{subfigure}%
\hfill %
    \begin{subfigure}[b]{0.325\textwidth}
        \centering
        \includegraphics[width=\linewidth]{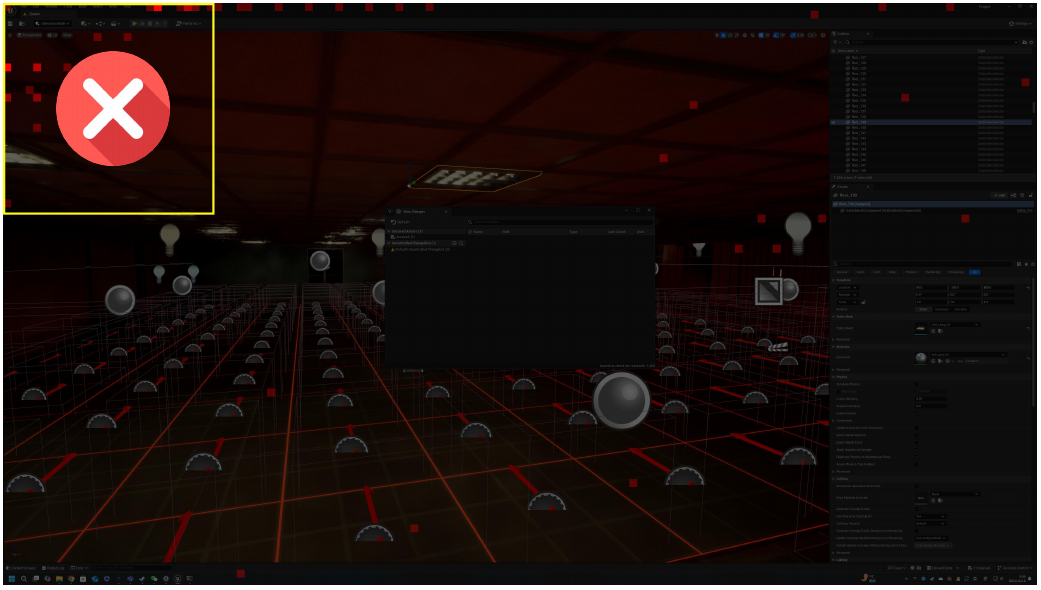}
        \caption{Prefill phase}
        \label{fig:unprefill}
    \end{subfigure}%
    \hfill %
    \begin{subfigure}[b]{0.325\textwidth}
        \centering
        \includegraphics[width=\linewidth]{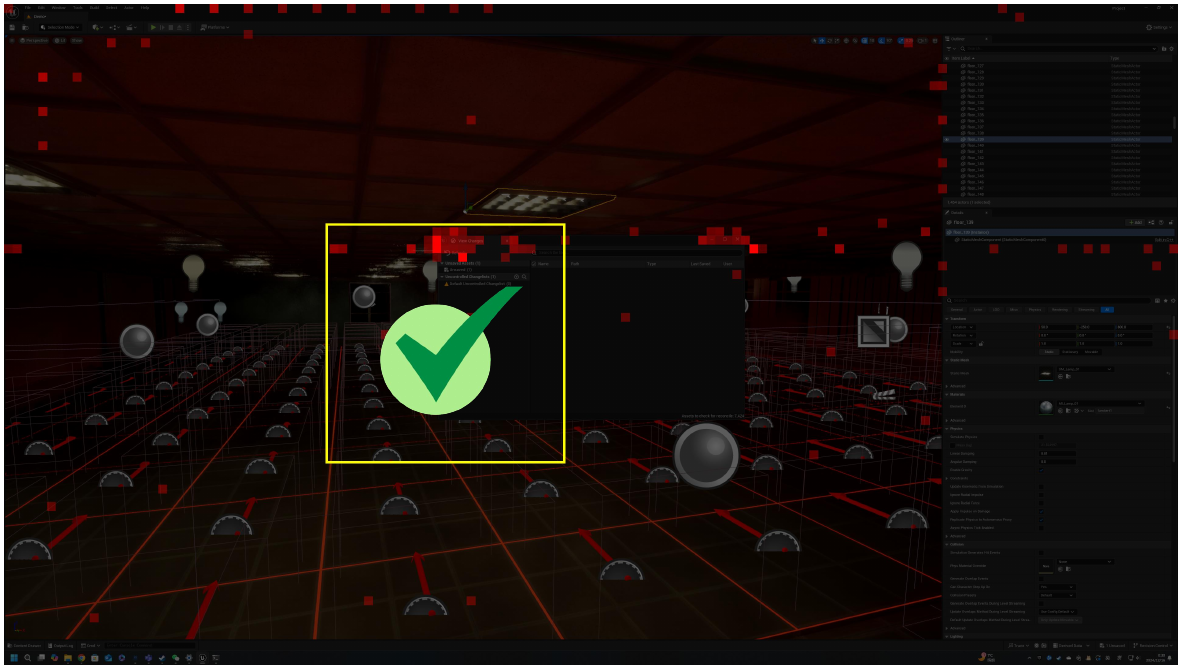}
        \caption{Generation phase}
        \label{fig:ungeneration}
    \end{subfigure}%
    \hfill %

    \caption{Visualization comparison. (a) is a UI interface with instruction: ``Refreshes changelists from revision control provider'', red block represents the ground truth region. Yellow blocks in attention maps of (b) and (c) are zoom-in regions that refer to the highest attention scores.}
    
    \label{fig:uncaptureatten}

\end{figure*}

\section{Visualizations of Attention Capture Phase}
\label{sec:vis2}

As discussed in main paper, the stage of attention capturing is a critical factor in determining the zoom-in region. In this section, we provide additional visualizations to further analyze the phase of capturing attention by comparing the attention distribution generated during the prefill and generation stages. We choose Web, iOS, and software interfaces with different resolutions, and leverage ZoomUI-7B to obtain these results in Figure~\ref{fig:oscaptureatten}, \ref{fig:ioscaptureatten}, and \ref{fig:uncaptureatten}. 

As shown in the visualizations, the MLLM exhibits severe attention confusion during the prefill phase rather than localizing target elements. We observe that the model indiscriminately engages in cross-modal alignment between the instruction and the interface, failing to awaken its spatial reasoning ability. Without a robust comprehension of how textual intents map to specific spatial regions, the attention mechanism becomes disoriented. Notably, we observe a consistent bias where prefill attention prefers to the interface edges, particularly the top-left corner. We attribute this to how MLLMs process images: visual tokens are read sequentially from top-left to bottom-right. Before generating an answer, MLLM's attention naturally rests on the first part of tokens by default. This bias is corrected during the subsequent generation phase. Once the model begins predicting the coordinates, its attention shifts away from the top-left corner and accurately focuses on target element region.

\begin{table}[t!]
\centering
\caption{Performance evaluation on the \textbf{ScreenSpot-Pro} benchmark using alternative MLLM backbones. The top block provides results in main paper for reference. The bottom block details performance of advanced MLLMs. \textbf{Bold} indicates the best result.}

\resizebox{0.99\textwidth}{!}{
    \setlength{\tabcolsep}{3pt} 
    \begin{tabular}{l | c | cc cc cc cc cc cc | c}
    \toprule
    \multirow{2}{*}{\textbf{Model}} & \multirow{2}{*}{\makecell{\textbf{Data}\\\textbf{Size}}} & \multicolumn{2}{c}{\textbf{CAD}} & \multicolumn{2}{c}{\textbf{Dev.}} & \multicolumn{2}{c}{\textbf{Creative}} & \multicolumn{2}{c}{\textbf{Scientific}} & \multicolumn{2}{c}{\textbf{Office}} & \multicolumn{2}{c|}{\textbf{OS}} & \multirow{2}{*}{\textbf{Avg.}} \\
    \cmidrule(lr){3-4} \cmidrule(lr){5-6} \cmidrule(lr){7-8} \cmidrule(lr){9-10} \cmidrule(lr){11-12} \cmidrule(lr){13-14}
     &  & Text & Icon & Text & Icon & Text & Icon & Text & Icon & Text & Icon & Text & Icon & \\
    \midrule
    \rowcolor{gray!20} 
    \multicolumn{15}{l}{\textit{Reference Results}}\\
    \midrule
    \textbf{\name-3B} & 0 & 47.7 & 20.3 & 52.6 & 18.6 & 45.5 & 14.0 & 52.8 & 25.5 & 62.2 & 37.7 & 54.2 & 22.5  & 40.2 \\
    \midrule
    \textbf{\name-7B} &  0 & 59.9 & 25.0 & 64.9 & 30.3 & 56.6 & 23.1 & 67.4 & 31.8 & 81.4 & 43.4 & 67.3 & 34.8 & 52.8 \\
    \midrule
    \rowcolor{gray!20}
    \multicolumn{15}{l}{\textit{Alternative Backbones}}\\
    \midrule
    \textbf{\name-4B} & 0 & 63.9 & 25.0 & \textbf{70.8} & 34.5 & \textbf{62.1} & 23.1 & \textbf{74.3} & \textbf{39.1} & 76.8 & 47.1 & \textbf{71.0} & \textbf{46.1}  & 55.9 \\
    \midrule
    \textbf{\name-8B} &  0 & \textbf{69.0} & \textbf{26.6} & 66.2 & \textbf{35.9} & 59.6 & \textbf{26.6} & 72.9 & 37.3 & \textbf{82.5} & \textbf{56.6} & 66.4 & 38.2  & \textbf{56.3} \\
    \bottomrule
    \end{tabular}
}
\label{ssproappendix}
\end{table}

\begin{table}[t!]
\centering
\caption{Performance evaluation on the \textbf{OSWorld-G} benchmark using alternative MLLM backbones. \textbf{Bold} indicates the best result.}
\resizebox{0.84\textwidth}{!}{
    \setlength{\tabcolsep}{5pt}
    \begin{tabular}{l | c | ccccc | c}
    \toprule
    \textbf{Model} & \textbf{Data Size} & \textbf{Text} & \textbf{Elem} & \textbf{Layout} & \textbf{Manip} & \textbf{Refuse} & \textbf{Avg.} \\
    \midrule
    \rowcolor{gray!20} 
    \multicolumn{8}{l}{\textit{Reference Results}}\\
    \midrule
     \textbf{\name-3B}  & 0 & 53.3 & 46.4 & 44.3 & 36.4 & 0.4 & 49.9 \\
     \textbf{\name-7B}  & 0 & 57.9 & 55.2 & 55.7 & 43.0 & 1.1 & 54.2  \\
    \midrule
    \rowcolor{gray!20} 
    \multicolumn{8}{l}{\textit{Alternative Backbones}}\\
     \textbf{\name-4B}  & 0 & 66.4 & 57.9 & 61.7 & 48.3 & 5.6 & 58.3 \\
     \textbf{\name-8B}  & 0 & \textbf{70.8} & \textbf{68.6} & \textbf{66.1} & \textbf{52.3} & \textbf{0} & \textbf{61.1}  \\
    \bottomrule
    \end{tabular}
}
\label{osworldappendix}

\end{table}

\begin{table}[t!]
\centering
\caption{Performance evaluation on the \textbf{UI-Vision} benchmark using alternative MLLM backbones. \textbf{Bold} indicates the best result.}
\resizebox{0.98\textwidth}{!}{
    \setlength{\tabcolsep}{2pt}
    \begin{tabular}{l | c | cccccc | ccc | c}
    \toprule
    \multirow{2}{*}{\textbf{Model}} & \multirow{2}{*}{\makecell{\textbf{Data}\\\textbf{Size}}} & \multicolumn{6}{c}{\textbf{Grouped by Category}} & \multicolumn{3}{c}{\textbf{Grouped by Setting}} & \multirow{2}{*}{\textbf{Avg.}} \\
    \cmidrule(lr){3-8} \cmidrule(lr){9-11}
     & & Edu. & Browser & Dev. & Prod. & Creative & Entert. & Basic & Func. & Spatial & \\
    \midrule
    \rowcolor{gray!20} 
    \multicolumn{12}{l}{\textit{Reference Results}}\\
    \midrule
     \textbf{\name-3B}  & 0 & 23.0 & 33.1 & 18.9 & 19.0 & 13.2 & 40.0 & 24.5 & 19.1 & 6.7 & 18.5 \\
     \textbf{\name-7B}  & 0 & 33.2 & 43.7 & 31.4 & 27.9 & 19.0 & \textbf{49.3} & 33.3 & 29.1 & 14.0 & 27.1 \\
    \midrule
    \rowcolor{gray!20} 
    \multicolumn{12}{l}{\textit{Alternative Backbones}}\\
     \textbf{\name-4B}  & 0 & 37.6 & 44.9 & 31.3 & \textbf{31.0} & 20.9 & 46.7 & 35.7 & 31.1 & \textbf{14.1} & 29.1 \\
     \textbf{\name-8B}  & 0 & \textbf{39.8} & \textbf{47.7} & \textbf{32.3} & 30.8 & \textbf{22.4} & 48.0 & \textbf{36.2} & \textbf{33.9} & \textbf{14.1} & \textbf{30.1} \\
    \bottomrule
    \end{tabular}
}
\label{uivisionappendix}

\end{table}

\begin{table}[t!]
\centering
\caption{Performance evaluation on the \textbf{MMBench-GUI} benchmark using alternative MLLM backbones. \textbf{Bold} indicates the best result.}
\resizebox{0.98\textwidth}{!}{
    \setlength{\tabcolsep}{2.5pt} 
    \begin{tabular}{l | c | cc cc cc cc cc cc | c}
    \toprule
    \multirow{2}{*}{\textbf{Model}} & \multirow{2}{*}{\makecell{\textbf{Data}\\\textbf{Size}}} & \multicolumn{2}{c}{\textbf{Windows}} & \multicolumn{2}{c}{\textbf{MacOS}} & \multicolumn{2}{c}{\textbf{Linux}} & \multicolumn{2}{c}{\textbf{iOS}} & \multicolumn{2}{c}{\textbf{Android}} & \multicolumn{2}{c|}{\textbf{Web}} & \multirow{2}{*}{\textbf{Avg.}} \\
    \cmidrule(lr){3-4} \cmidrule(lr){5-6} \cmidrule(lr){7-8} \cmidrule(lr){9-10} \cmidrule(lr){11-12} \cmidrule(lr){13-14}
     & & Bas. & Adv. & Bas. & Adv. & Bas. & Adv. & Bas. & Adv. & Bas. & Adv. & Bas. & Adv. & \\
    \midrule
    \rowcolor{gray!20} 
    \multicolumn{15}{l}{\textit{Reference Results}}\\
    \textbf{\name-3B} & 0 & 65.1 & 43.1 & 71.4 & 48.0 & 75.9 & 38.3 & 77.9 & 70.4 & 76.9 & 70.5 & 65.0 & 62.9 & 63.2 \\
    \textbf{\name-7B} & 0 & 73.6 & 49.6 & 77.6 & 60.8 & 78.5 & 47.7 & 75.3 & 78.2 & 76.6 & 79.9 & 75.4 & 69.9 & 72.8 \\
    \midrule
    \rowcolor{gray!20} 
    \multicolumn{15}{l}{\textit{Alternative Backbones}}\\
    \midrule
    \textbf{\name-4B} & 0 & \textbf{77.0} & \textbf{65.7} & 77.9 & \textbf{61.4} & 82.3 & 54.5 & 81.5 & 83.4 & 83.8 & \textbf{83.9} & \textbf{82.1} & 72.4 & 75.4 \\
    \textbf{\name-8B} & 0 & 75.8 & 61.3 & \textbf{79.6} & \textbf{61.4} & \textbf{94.9} & \textbf{57.5} & \textbf{84.2} & \textbf{85.2} & \textbf{85.1} & 83.6 & 76.0 & \textbf{74.6} & \textbf{75.7} \\
    \bottomrule
    \end{tabular}
    \label{mmbenchappendix}
}
\end{table}

\section{Performance Evaluation on Alternative MLLMs}
\label{sec:othermllm}
As discussed in the limitations of main paper, our primary evaluations are conducted using the Qwen2.5-VL architectures. To further validate our proposed method, we extend evaluations to the newly released Qwen3-VL series in this section. Specifically, we integrate our method with the Qwen3-VL-4B and Qwen3-VL-8B base models, denoting the GUI agents as ZoomUI-4B and ZoomUI-8B, respectively. We evaluate them on four benchmarks in Table~\ref{ssproappendix}, \ref{osworldappendix}, \ref{uivisionappendix} and~\ref{mmbenchappendix}.

The experimental results demonstrate that performance of ZoomUI is further enhanced when integrated with more advanced MLLM backbones. Notably, ZoomUI-8B significantly outperforms several fine-tuning based GUI agents reported in the main paper despite being completely train-free. This advancement further validates that a proper inference scaling strategy can unlock the inherent capability of common MLLMs, achieving superior GUI grounding without any reliance on specialized data fine-tuning. Moreover, these findings suggest a promising future direction: targeted introduction of data adaptation during the inference stage could potentially help common MLLMs further evolve, thereby unlocking more powerful and precise GUI grounding performance.

\end{document}